\PassOptionsToPackage{table}{xcolor}
\documentclass{article} 
\usepackage{iclr2027_conference_arxiv,times}
\iclrfinalcopy

\usepackage{amsmath,amsfonts,bm}

\def\eqref#1{equation~\ref{#1}}

\def\1{\bm{1}}

\DeclareMathAlphabet{\mathsfit}{\encodingdefault}{\sfdefault}{m}{sl}
\SetMathAlphabet{\mathsfit}{bold}{\encodingdefault}{\sfdefault}{bx}{n}

\usepackage{hyperref}
\usepackage{url}

\newcommand{\method}{TQG-Exp}

\usepackage{algorithm,algorithmic}
\usepackage{hyperref}
\usepackage{multirow}
\usepackage{subcaption}  

\usepackage{pifont}

\newcommand{\xmark}{\ding{55}}
\usepackage{booktabs,tabularx}

\PassOptionsToPackage{hyphens}{url}
\PassOptionsToPackage{breaklinks,colorlinks}{hyperref}

\usepackage{graphicx} 
\usepackage{xurl}              
\usepackage{hyperref}          
\usepackage[table]{xcolor} 
\usepackage{array}
\newcolumntype{R}{>{\color{red}}c}
\newcommand{\Src}[1]{\href{#1}{\url{#1}}}

\usepackage{tcolorbox}
\usepackage{xcolor}
\tcbuselibrary{skins,breakable}
\usepackage{amsthm}  
\usepackage{wrapfig}

\newtcolorbox{promptbox}[1]{
  enhanced,
  breakable,
  colback=gray!2,
  colframe=gray!30,
  colbacktitle=gray!10,
  coltitle=black,
  boxrule=0.5pt,
  arc=1pt,
  left=7pt,
  right=7pt,
  top=6pt,
  bottom=6pt,
  before skip=8pt,
  after skip=8pt,
  title=#1,
  fonttitle=\bfseries\small,
  fontupper=\small,
  titlerule=0pt
}

\usepackage{listings}

\usepackage{listings}
\usepackage{xcolor}

\definecolor{jsonbg}{RGB}{248,249,251}
\definecolor{jsonframe}{RGB}{210,214,220}
\definecolor{jsonkey}{RGB}{45,85,145}
\definecolor{jsonstring}{RGB}{45,110,75}
\definecolor{jsonnumber}{RGB}{155,80,45}

\lstdefinestyle{jsonstyle}{
    basicstyle=\ttfamily\scriptsize,
    backgroundcolor=\color{jsonbg},
    frame=single,
    rulecolor=\color{jsonframe},
    framerule=0.5pt,
    framesep=6pt,
    xleftmargin=0.6em,
    xrightmargin=0.6em,
    aboveskip=0.8em,
    belowskip=0.8em,
    breaklines=true,
    breakatwhitespace=false,
    columns=fullflexible,
    keepspaces=true,
    showstringspaces=false,
    numbers=none,
    tabsize=2,
    upquote=true,
    stringstyle=\color{jsonstring},
    literate=
      *{0}{{{\color{jsonnumber}0}}}{1}
       {1}{{{\color{jsonnumber}1}}}{1}
       {2}{{{\color{jsonnumber}2}}}{1}
       {3}{{{\color{jsonnumber}3}}}{1}
       {4}{{{\color{jsonnumber}4}}}{1}
       {5}{{{\color{jsonnumber}5}}}{1}
       {6}{{{\color{jsonnumber}6}}}{1}
       {7}{{{\color{jsonnumber}7}}}{1}
       {8}{{{\color{jsonnumber}8}}}{1}
       {9}{{{\color{jsonnumber}9}}}{1}
}
\title{Beyond Outcome Feedback: Trajectory-Quality-Guided Experience-based Learning}

\author{ \hspace*{-0.25em}\textbf{Bingjie Zhang}$^{1}$, \: \textbf{Yibo Yang}$^{2}$\thanks{Corresponding author.}, \: \textbf{Qiantong Liu}$^{1}$, \: \textbf{Benshuo Fu}$^{1}$, \: \textbf{ Yi Chang}$^{1}$, \: \textbf{Dandan Guo}$^{1,3}$\footnotemark[1]\\[0.0em] 
\begin{tabular}[t]{@{}l@{}} 
$^{1}$School of Artificial Intelligence, Jilin University\\ $^{2}$School of Artificial Intelligence, Shanghai Jiao Tong University \\
$^{3}$King Abdullah University of Science and Technology 
\\ \texttt{zhangbj24@mails.jlu.edu.cn}, \: \texttt{yibo.yang@sjtu.edu.cn}, \\ \texttt{guodandan@jlu.edu.cn} \end{tabular} }
\begin{document}

\maketitle

\begin{abstract}
Experience-based learning enables large language model (LLM) agents to improve over time by reusing knowledge from past executions. However, existing methods typically rely on final task outcomes as feedback for experience learning, overlooking the execution process and limiting their ability to evaluate and optimize experience based on trajectory quality. To address this limitation, we propose \method, a trajectory-quality-guided experience-based learning framework that introduces process feedback. Specifically, we introduce Trajectory Quality Loss (TQL), which jointly considers task completion and process quality to provide a learning signal beyond final task outcomes. Based on this signal, we employ paired rollouts with and
without retrieved experience to guide experience optimization. Representative trajectories are selected based on their outcomes and trajectory quality to extract task-level and task-type-level experiences. Meanwhile, differences in trajectory quality between the paired groups are used to estimate the contribution
of retrieved experience and update its utility. The learned utility is then combined with semantic  relevance to guide subsequent experience retrieval. Experiments on ALFWorld, WebShop, and ScienceWorld with two backbone LLMs show that \method~ outperforms existing experience-based methods, improving Avg.@3 over the strongest baseline by 4.50 and 8.55 percentage points on DeepSeek-V4-Flash and Qwen3.5-Flash, respectively.
\end{abstract}    

\section{Introduction}

\begin{figure}[h]
    \centering
    \vspace{-1.5em}
    \includegraphics[
        width=\linewidth,
        trim=2pt 2pt 2pt 2pt,
        clip
    ]{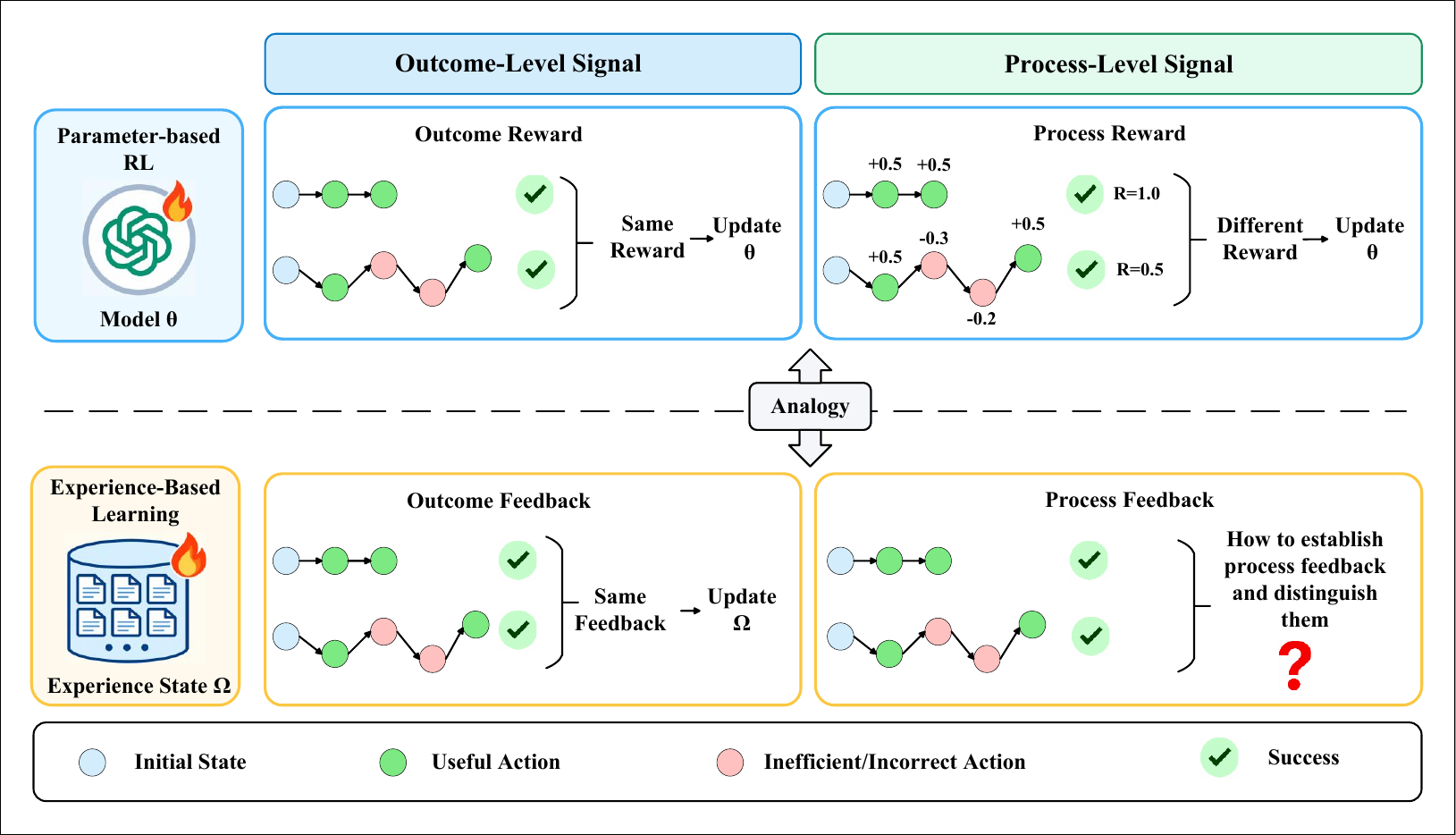}
    \caption{Illustration of outcome-level and process-level signal in reinforcement learning (top) and experience-based learning (bottom).}
    \label{fig:representation_adaptation}
\end{figure}

Recent advances in large language models (LLMs) have substantially expanded the capabilities of LLM agents, enabling them to tackle complex interactive tasks such as embodied decision making, web navigation, and scientific exploration \citep{shinn2023reflexion, tan2026coarse, team2025tongyi}. Despite these advances, many LLM agents still operate in an episodic manner, with each task typically solved independently of previous executions \citep{zhang2025memevolve}. Consequently, useful information contained in successful and failed executions is often lost once an episode ends, forcing agents to rediscover effective strategies or repeat similar mistakes in subsequent tasks \citep{yin2025learning,gao2025survey}. 
This limitation has motivated growing interest in experience-based learning, which enables agents to leverage experience from past executions to improve over time and transfer useful knowledge across tasks.

Existing experience-based learning methods typically evaluate experience based on final task outcomes, such as success or failure \citep{zhang2026memrl, tu2026dynamic, feng2026expgraph, tan2026deltamem}. However, task success does not necessarily imply high-quality experience, and task failure does not necessarily mean that the experience is entirely useless. This is because final outcomes cannot distinguish trajectories that share the same result but differ substantially in process quality. A successful trajectory may still contain redundant or invalid actions and unnecessarily long execution steps, while a failed trajectory may make substantial progress and fail only near the end. In other words, binary success or failure labels cannot adequately reflect intermediate progress and execution efficiency.

As shown in Fig.~\ref{fig:representation_adaptation}, experience-based learning can be viewed in analogy to parameter-based reinforcement learning: both use learning signals to optimize an adaptable component \citep{zhang2026memrl}. In conventional RL, reward signals guide policy optimization, typically through parameter updates. In experience-based learning, the model parameters remain fixed, while outcome feedback is used to update the external experience, determining which experiences are retained, assigned value, and reused in later decisions.  In parameter-based RL, relying only on outcome rewards provides a relatively coarse learning signal and cannot distinguish different execution processes that lead to the same result. The same issue arises in experience-based learning, where the final outcome may not accurately reflect how an experience affects the execution process.  Currently, some works in reinforcement learning explore process rewards to provide richer supervision for policy optimization \citep{wang2024math,setlur2025rewarding,chen2025discriminative}. Following the same idea, experience learning also calls for process feedback to better evaluate and optimize external experience. This raises a key question: How can we establish process feedback to distinguish trajectory quality beyond final task outcomes?

To this end, we propose \method,  Trajectory-Quality-Guided
Experience-Based Learning, which introduces process feedback
into experience learning. We first introduce Trajectory Quality Loss (TQL), a process signal that jointly considers task completion and  process quality, allowing trajectories with the same outcome to be distinguished by how well they are executed. We then use this signal to guide external experience optimization. Specifically, guided by TQL, paired rollouts with and without retrieved experience are used to select representative trajectories for extracting task- and task-type-level experiences and to estimate the contribution of retrieved experience based on differences in trajectory quality. The estimated utility is further combined with semantic relevance to guide subsequent experience retrieval, forming a closed loop of experience extraction, evaluation, and reuse.

In summary, our contributions are threefold:

\noindent\textbullet\ We formulate experience-based learning from a trajectory-quality perspective, moving beyond outcome-only feedback toward process-aware supervision for evaluating and optimizing external experience while keeping the LLM fixed.

\noindent\textbullet\  We propose \method, a trajectory-quality-guided experience-based learning framework. It introduces Trajectory Quality Loss (TQL) to evaluate both task completion and process quality. Guided by TQL, paired rollouts with and without retrieved experience are used to select representative trajectories for extracting task- and task-type-level experiences and to estimate the incremental contribution of retrieved experience. The estimated utility is then combined with semantic relevance to guide subsequent experience retrieval.

\noindent\textbullet\ We evaluate \method~on ALFWorld, WebShop, and ScienceWorld with two base models. \method~consistently outperforms existing experience-based methods, improving Avg.@3 over the strongest baseline by 4.50 percentage points on DeepSeek-V4-Flash and 8.55 percentage points on Qwen3.5-Flash.

\begin{figure}[H]

    \centering
  \includegraphics[
    width=\textwidth,
    trim=2pt 1.5pt 1.5pt 1.5pt,
    clip
]{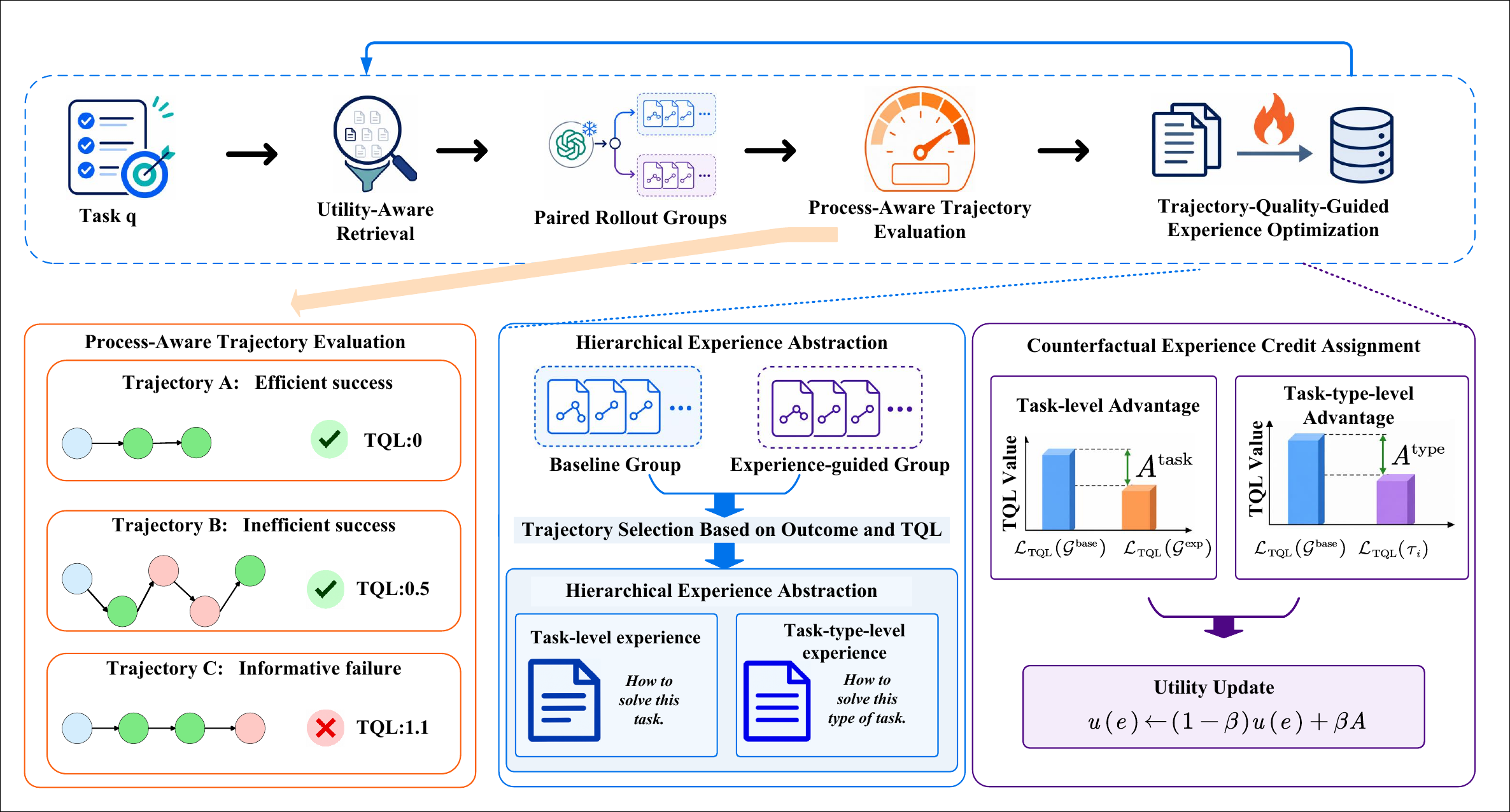}
    \caption{
    Overview of \method. The top illustrates the overall experience learning loop, while the bottom panels detail process-aware trajectory evaluation, hierarchical experience abstraction, and counterfactual experience credit assignment with utility updates.
    }
    \label{fig:overview}
\end{figure}

\section{Related Work}

\textbf{LLM Agents.} The rapid progress of LLMs has driven growing interest in autonomous agents that interact with external environments \citep{wang2024survey, wei2026agentic}. Representative frameworks such as ReAct enable agents to interleave reasoning with actions and adapt to environmental feedback \citep{yao2023react}, while AutoGen and CAMEL further support multi-agent collaboration and tool-augmented problem solving \citep{wu2024autogen,licamel}. Despite these advances, many agents still rely on transient prompt-level context, such as recent interaction histories or task-specific hints \citep{dong2024survey}. Such information is usually tied to the current episode and is not retained for future use. This is limiting in long-horizon settings, where tasks often share subgoals, action patterns, and failure modes \citep{zhao2024expel}. As a result, agents may repeatedly make similar mistakes or rediscover effective strategies from scratch \citep{xiong2025memory, wang2025mirix, hu2026continual}. These limitations have motivated experience-based learning, where agents extract reusable knowledge from past executions and use it to improve decision making on subsequent tasks.

\textbf{Experience-Based Learning for LLM Agents.}
Recent work has explored how LLM agents can improve from past interactions by extracting and reusing experience. Existing approaches mainly differ in whether experience learning is coupled with policy optimization or performed externally with a frozen LLM. SKILLRL \citep{xia2026skillrl} and D2Skill \citep{tu2026dynamic} jointly evolve the policy and reusable experience, while methods such as ReasoningBank \citep{ouyang2025reasoningbank}, MemP \citep{fang2025memp}, Training-Free GRPO \citep{cai2025training}, MemRL \citep{zhang2026memrl}, and DELTAMEM \citep{tan2026deltamem} keep the underlying LLM fixed and improve subsequent behavior through externally stored experience. Despite differences in how experience is represented and optimized, these methods largely rely on final task outcomes or outcome-derived signals to evaluate and refine experience. Such signals provide limited information about the quality of the execution process itself. Our work instead introduces trajectory quality as the supervision signal for experience learning, allowing experience to be evaluated beyond final outcomes.

\section{Preliminaries}

\textbf{LLM Agent.}  Following \cite{tu2026dynamic}, we adopt the ReAct framework
\citep{yao2022react} and formulate each task as a partially observable
sequential interaction process
$(\mathcal{S},\mathcal{A},\mathcal{O},\mathcal{T})$,
where $\mathcal{S}$, $\mathcal{A}$, and $\mathcal{O}$ denote the state,
action, and observation spaces, respectively, and
$\mathcal{T}:\mathcal{S}\times\mathcal{A}\rightarrow\mathcal{S}$
is the environment transition function. Since the underlying environment
state is not directly accessible, the agent makes decisions based on the
task query and its interaction history.

At step $t$, the agent observes the context
\begin{equation}
h_t=(q,o_1,a_1,o_2,\ldots,a_{t-1},o_t),
\end{equation}
where $q$ denotes the task query, $a_i\in\mathcal{A}$ is the action taken
at step $i$, and $o_i\in\mathcal{O}$ is the corresponding observation.
Conditioned on $h_t$, the frozen LLM policy $\pi_\theta$ generates:

\begin{equation}
R_t=\langle z_t,a_t\rangle \sim \pi_\theta(\cdot\mid h_t),
\end{equation}

where $z_t$ denotes the reasoning trace and $a_t$ the action to execute.
After $a_t$ is executed, the environment returns the next observation
$o_{t+1}$. The reasoning trace $z_t$ is discarded, while
$(a_t,o_{t+1})$ is appended to the interaction history for the next step. The interaction terminates when the agent produces a final answer $y$ or
reaches the maximum number of steps. The resulting trajectory is:
\begin{equation}
\tau=(q,o_1,a_1,o_2,\ldots,a_{T-1},o_T,y).
\end{equation}

\textbf{Experience-Augmented Agent.}
An experience-augmented agent extends the interaction context with external experience retrieved from an experience space $\mathcal{E}$. Given the current query and interaction history, a retrieval function selects a set of relevant experiences: \begin{equation} \mathcal{E}_t^{\mathrm{ret}} = \operatorname{Retrieve}(q,h_t,\mathcal{E}). \label{3-2-1} \end{equation} The retrieved experience is incorporated into the agent context, and the response is generated as: \begin{equation} R_t = \langle z_t,a_t\rangle \sim \pi_{\theta} \left( \cdot \mid h_t,\mathcal{E}_t^{\mathrm{ret}} \right), \label{3-2-2} \end{equation} where $\pi_{\theta}$ denotes the frozen LLM policy. The experience space can be constructed from previous trajectories or distilled experiences and provides additional context for action selection without modifying the LLM parameters.


\section{Method}


To overcome the limitations of outcome-based feedback in experience learning, we propose \method, a new approach that integrates process-aware trajectory evaluation with trajectory-quality-guided experience optimization. As illustrated in Fig.~\ref{fig:overview}, we introduce Trajectory Quality Loss (TQL) to provide process feedback beyond final task outcomes.
Based on this signal, we employ paired rollouts
with and without retrieved experience to select representative trajectories for experience extraction
and estimate experience utility based on differences in trajectory quality. The learned utility then
guides subsequent experience retrieval, forming a closed loop of experience optimization and reuse.

\subsection{Process-Aware Trajectory Evaluation}
We first focus on constructing a process-aware learning signal for experience learning. Process rewards in reinforcement learning provide finer-grained reward signals beyond final outcome rewards. Following this idea, we introduce process feedback into experience learning without relying on a learned reward model. Instead, we directly construct a trajectory-level quality signal from the generated execution, termed Trajectory Quality Loss (TQL).

\textbf{Trajectory Quality Loss.}
For a trajectory $\tau_i$, we define:
\begin{equation}
\label{eq:tql}
\mathcal{L}_{\mathrm{TQL}}(\tau_i)
=
(1-y_i)
+
\left(1-\mathrm{PQ}(\tau_i)\right),
\end{equation}
where $y_i\in\{0,1\}$ indicates task completion and
$\mathrm{PQ}(\tau_i)$ measures process quality based on trajectory length,
invalid actions, repeated-action loops, and the position of the first error. These signals characterize different aspects of the execution process, including efficiency,  executability, behavioral stability, and  progress before failure. Lower TQL  indicates higher trajectory quality. The detailed definition of
$\mathrm{PQ}(\tau_i)$ is provided in Appendix~\ref{sec:tql}.

\textbf{Group-Level Trajectory Quality.} For each task $q$, multiple rollouts form a group $\mathcal{G}$. We define the group-level TQL as:
\begin{equation}
\label{eq:group_tql}
\mathcal{L}_{\mathrm{TQL}}(\mathcal{G})
=
\frac{1}{|\mathcal{G}|}
\sum_{\tau_i\in\mathcal{G}}
\mathcal{L}_{\mathrm{TQL}}(\tau_i)
+
\left(
1-
\frac{1}{|\mathcal{G}|}
\sum_{\tau_i\in\mathcal{G}} y_i
\right).
\end{equation}
The trajectory-level TQL evaluates the quality of an individual execution, whereas the group-level TQL summarizes the overall quality of multiple rollouts under the same task condition. In addition to the average trajectory-level TQL, the group-level formulation incorporates the group failure rate, which captures how frequently the agent fails across multiple rollouts under the same task condition.

\subsection{Trajectory-Quality-Guided Experience Optimization}
\label{sec:Traj_Qua_Gui}
Having constructed the trajectory-quality signal, we next use it to optimize
the external experience. We represent
the adaptable experience state as:
\begin{equation}
\Omega
=
\left(
\mathcal{E},
\mathcal{U},
\mathcal{N}
\right),
\end{equation}
where $\mathcal{E}$ denotes the experience space composed of stored experiences $e$, and $\mathcal{U}=\{u(e)\mid e\in\mathcal{E}\}$ and $\mathcal{N}=\{n(e)\mid e\in\mathcal{E}\}$ denote the utility values and retrieval counts associated with these experiences, respectively. Experience-based learning then aims to optimize this external state so that the
frozen LLM produces higher-quality trajectories:
\begin{equation}
\label{eq:external_objective}
\Omega^{\star}
=
\underset{\Omega}{\arg\min}\;
\mathbb{E}_{q\sim\mathcal{D},\,
\tau\sim\pi_{\theta,\Omega}(\cdot\mid q)}
\left[
\mathcal{L}_{\mathrm{TQL}}(\tau)
\right],
\end{equation}
where $\theta$ remains fixed, $\pi_{\theta,\Omega}$ denotes the policy induced by the frozen LLM conditioned on the external experience state $\Omega$, $\mathcal{D}$ denotes the task distribution, and
$\mathcal{L}_{\mathrm{TQL}}(\tau)$ measures trajectory quality.  We optimize the external experience state $\Omega$ through grouped rollouts under each task condition. For each task $q$, we perform paired rollouts consisting of a baseline group $\mathcal{G}^{\mathrm{base}}$ without retrieved experience and an experience-guided group $\mathcal{G}^{\mathrm{exp}}$ conditioned on retrieved experience. These rollout groups serve two purposes. First, trajectory quality is used to select representative trajectories from the groups for updating the experience space $\mathcal{E}$. Second, the relative quality of the paired groups is used to estimate the contribution of retrieved experience and update its utility $\mathcal{U}$. The learned utility, together with retrieval statistics $\mathcal{N}$, then guides subsequent experience retrieval. This process forms a closed loop of experience extraction, evaluation, and reuse.


\textbf{Quality-Guided Hierarchical Experience Abstraction.}
Since the learned experience is intended to address limitations revealed during agent execution, we use baseline rollouts as the primary source of failure cases. Representative trajectories are then selected according to task outcome and trajectory-level TQL, favoring high-quality successes and contrasting them with informative failures or inefficient successes from the baseline group. The selected trajectories are distilled into two levels of experience: task-level experience captures task-specific procedures, key considerations, and failure patterns, while task-type-level experience summarizes reusable decision patterns across related tasks. Detailed extraction rules and the extraction prompt are provided in Appendix~\ref{sec:exp_extra} and Appendix~\ref{sec:prompt_and_Exam}, respectively.

\textbf{Counterfactual Experience Credit Assignment.} A single TQL value cannot determine whether retrieved experience is genuinely
effective, since a successful trajectory may result  from the agent's inherent capabilities rather than from the retrieved experience itself.
We therefore compare paired rollouts $\mathcal{G}^{\mathrm{base}}$ and
$\mathcal{G}^{\mathrm{exp}}$ under the same task $q$ to estimate the
incremental contribution of retrieved experience.
Task-level experience is retrieved once for each task and shared across its
rollouts, whereas task-type-level experience is retrieved dynamically at each
interaction step and may therefore differ across trajectories. We accordingly
estimate their contributions at different granularities. For task-level
experience, the contribution is measured by the difference in group-level TQL:

\begin{equation}
\label{eq:task_advantage}
A^{\mathrm{task}}(q)
=
\mathcal{L}_{\mathrm{TQL}}
\left(\mathcal{G}^{\mathrm{base}}\right)
-
\mathcal{L}_{\mathrm{TQL}}
\left(\mathcal{G}^{\mathrm{exp}}\right).
\end{equation}

For task-type-level experience, we evaluate each guided trajectory relative
to the overall baseline performance:
\begin{equation}
\label{eq:type_advantage}
A^{\mathrm{type}}(\tau_i\mid q)
=
\mathcal{L}_{\mathrm{TQL}}
\left(\mathcal{G}^{\mathrm{base}}\right)
-
\mathcal{L}_{\mathrm{TQL}}(\tau_i),
\qquad
\tau_i\in\mathcal{G}^{\mathrm{exp}}.
\end{equation}

The group-level and trajectory-level TQL values are normalized to $[0,1]$ before computing the advantages. The resulting advantages are then used to update experience utility:
\begin{equation}
u^{l}(e)
\leftarrow
(1-\beta_l)u^{l}(e)
+
\beta_l A^{l},
\qquad
l\in\{\mathrm{task},\mathrm{type}\}.
\end{equation}
 A positive advantage indicates improvement over the baseline reference.

\textbf{Utility-Aware Experience Retrieval.} 
For subsequent tasks, we retrieve candidates
$\mathcal{C}(q)$ based on semantic relevance,
restricted to the same task type and excluding
the current task's own experiences.
Each candidate $e \in \mathcal{C}(q)$ is then
reranked using semantic relevance and learned utility:
\begin{equation}
\label{eq:routing}
\mathcal{R}_{K}(q)
=
\operatorname{TopK}_{e\in\mathcal{C}(q)}
\left[
\alpha\,\mathrm{sim}(q,e)
+
(1-\alpha)\left[
u(e)
+
c\sqrt{
\frac{\log(1+N)}
{1+n(e)}
}
\right]
\right],
\end{equation}
where $N$ is the total retrieval count, $\alpha$ is the relevance-utility trade-off coefficient, and $c$ controls exploration. The top-$K$ experiences under this combined score are
selected for subsequent use. During evaluation, we also retrieve successful trajectories
from the top-$K_{\mathrm{tra}}$ most similar tasks as additional
execution exemplars. Details are
provided in Appendix~\ref{sec:trajectory_retrieval}.


\section{Experiments}

\subsection{Setup}
\label{sec:setup}
\textbf{Datasets.} Referring to the benchmark settings used in related studies \citep{yang2025coarse, shang2025agentsquare, zhang2026memrl, tu2026dynamic,tan2026deltamem}, we evaluate \method~on three challenging benchmarks for LLM agents: ALFWorld \citep{shridhar2020alfworld}, WebShop \citep{yao2022webshop}, and ScienceWorld \citep{wang2022scienceworld}. Further details of the datasets are provided in Appendix~\ref{sec:data_setting}.

\textbf{Base models.} We conduct experiments with DeepSeek-V4-Flash  and Qwen3.5-Flash. For each model, the same LLM serves as both the agent model for environment interaction and the reflection model for extracting experiences from the selected trajectories, without requiring an additional LLM.

\textbf{Baseline Methods.}
We compare our method with five baselines: No Experience, ReasoningBank \citep{ouyang2025reasoningbank}, MemP \citep{fang2025memp}, MemRL \citep{zhang2026memrl}, and DeltaMem \citep{tan2026deltamem}. Implementation details and configuration settings are provided in Appendix~\ref{sec:baseline_sum}.

\textbf{Evaluation Metrics.} We report Average@3 and Pass@3. Average@3 is the average success rate across three independent rollouts, while Pass@3 is the percentage of tasks solved in at least one of the three rollouts.

\subsection{Main results}

\begin{table}[H]
\centering
\caption{
Performance comparison across ALFWorld, WebShop, and ScienceWorld.
Best results are in \textbf{bold}, and second-best results are
\underline{underlined}.
}
\label{tab:main_results}
\setlength{\tabcolsep}{3.8pt}
\renewcommand{\arraystretch}{1.10}

\resizebox{\textwidth}{!}{
\begin{tabular}{llcccccccccc}
\toprule

\multirow{3}{*}{\textbf{Model}}
&
\multirow{3}{*}{\textbf{Method}}
&
\multicolumn{4}{c}{\textbf{ALFWorld}}
&
\multicolumn{2}{c}{\multirow{2}{*}{\textbf{WebShop}}}
&
\multicolumn{2}{c}{\multirow{2}{*}{\textbf{ScienceWorld}}}
&
\multicolumn{2}{c}{\multirow{2}{*}{\textbf{Avg.}}}
\\

\cmidrule(lr){3-6}

& &
\multicolumn{2}{c}{\textbf{Seen}}
&
\multicolumn{2}{c}{\textbf{Unseen}}
&
\multicolumn{2}{c}{}
&
\multicolumn{2}{c}{}
&
\multicolumn{2}{c}{}
\\

\cmidrule(lr){3-4}
\cmidrule(lr){5-6}
\cmidrule(lr){7-8}
\cmidrule(lr){9-10}
\cmidrule(lr){11-12}

& &
\textbf{Avg.@3} $\uparrow$
& \textbf{Pass@3} $\uparrow$
& \textbf{Avg.@3} $\uparrow$
& \textbf{Pass@3} $\uparrow$
& \textbf{Avg.@3} $\uparrow$
& \textbf{Pass@3} $\uparrow$
& \textbf{Avg.@3} $\uparrow$
& \textbf{Pass@3} $\uparrow$
& \textbf{Avg.@3} $\uparrow$
& \textbf{Pass@3} $\uparrow$
\\

\midrule

\multirow{6}{*}{\textbf{DeepSeek-V4-Flash}}
& No Experience
& 68.57 & 76.43
& 74.88 & 85.82
& 42.33 & 52.00
& 47.00 & 60.00
& 58.20 & 68.56
\\

& ReasoningBank \citep{ouyang2025reasoningbank}
& 75.24 & 82.14
& 75.37 & 85.07
& 39.67 & 48.00
& 48.33 & 55.00
& 59.65 & 67.55
\\

& MemP \citep{fang2025memp}
& 84.76 & 91.43
& 88.06 & \underline{94.03}
& 36.33 & 48.00
& 57.33 & 63.00
& 66.62 & 74.12
\\

& MemRL \citep{zhang2026memrl}
& \underline{90.00} & \underline{92.86}
& \underline{90.30} & 93.28
& \underline{73.67} & \underline{78.00}
& \underline{69.33} & \underline{76.00}
& \underline{80.83} & \underline{85.04}
\\

& DELTAMEM \citep{tan2026deltamem}
& 80.95 & 87.14
& 79.85 & 85.82
& 48.33 & 60.00
& 56.67 & 65.00
& 66.45 & 74.49
\\

\rowcolor{gray!10}
& \textbf{\method}
& \textbf{96.19} & \textbf{96.43}
& \textbf{90.80} & \textbf{94.78}
& \textbf{79.67} & \textbf{85.00}
& \textbf{74.67} & \textbf{81.00}
& \textbf{85.33} & \textbf{89.30}
\\

\midrule

\multirow{6}{*}{\textbf{Qwen3.5-Flash}}
& No Experience
& 65.00 & 75.00
& 70.40 & 79.10
& 51.67 & 67.00
& 42.00 & 55.00
& 57.27 & 69.03
\\

& ReasoningBank \citep{ouyang2025reasoningbank}
& 73.10 & 80.00
& 79.60 & 88.06
& 49.33 & 61.00
& 46.67 & 57.00
& 62.18 & 71.52
\\

& MemP \citep{fang2025memp}
& 79.52 & 87.86
& 85.32 & \underline{91.04}
& 60.33 & 73.00
& \underline{58.67} & \underline{73.00}
& 70.96 & 81.23
\\

& MemRL \citep{zhang2026memrl}
& \underline{87.14} & \underline{89.29}
& \underline{85.82} & 89.55
& \underline{75.00} & \underline{79.00}
& 57.67 & 70.00
& \underline{76.41} & \underline{81.96}
\\

& DELTAMEM \citep{tan2026deltamem}
& 72.62 & 79.29
& 76.37 & 85.82
& 60.00 & 72.00
& 54.67 & 68.00
& 65.92 & 76.28
\\

\rowcolor{gray!10}
& \textbf{\method}
& \textbf{93.10} & \textbf{96.43}
& \textbf{89.05} & \textbf{93.28}
& \textbf{80.00} & \textbf{82.00}
& \textbf{77.67} & \textbf{94.00}
& \textbf{84.96} & \textbf{91.43}
\\

\bottomrule
\end{tabular}
}
\end{table}

Tab.~\ref{tab:main_results} compares \method~with existing experience-based methods across three interactive benchmarks and two backbone LLMs.
\method~achieves the best Average@3 and Pass@3 in all evaluated settings.
Among the baselines, MemRL achieves the strongest average performance across the evaluated settings. It stores complete trajectories as experiences and updates their utility through runtime reinforcement learning, enabling retrieval based on both semantic relevance and learned utility.
Nevertheless, \method~consistently surpasses MemRL with both backbone LLMs.
With DeepSeek-V4-Flash, the overall Average@3 increases from 80.83 to 85.33
and Pass@3 from 85.04 to 89.30. With Qwen3.5-Flash, Average@3 improves from
76.41 to 84.96 and Pass@3 from 81.96 to 91.43.

The advantage over MemRL is also consistent at the benchmark level. On
ALFWorld, \method~outperforms MemRL on both seen and unseen splits with both
backbone models, suggesting that the learned experience remains useful on
unseen task instances. The gains extend to WebShop and ScienceWorld. In
particular, Average@3 improves over MemRL by 6.00 points on WebShop with
DeepSeek-V4-Flash and by 20.00 points on ScienceWorld with Qwen3.5-Flash.
Taken together, these results highlight the value of process feedback in experience learning, where TQL provides a finer-grained learning signal beyond task outcomes.

\subsection{Performance and Interaction Efficiency}

\begin{wraptable}[11]{l}{0.60\linewidth}
\centering
\vspace{-1.3em}
\caption{Performance and efficiency on ALFWorld with DeepSeek-V4-Flash.}
\label{tab:alfworld_effi}
\vspace{-0.5em}
\small
\setlength{\tabcolsep}{3.2pt}
\renewcommand{\arraystretch}{1.03}

\resizebox{\linewidth}{!}{
\begin{tabular}{llcccc}
\toprule

\multirow{2}{*}{\textbf{Split}} &
\multirow{2}{*}{\textbf{Method}} &
\multicolumn{4}{c}{\textbf{Performance \& Efficiency}} \\
\cmidrule(lr){3-6}

& &
\textbf{Avg.@3} $\uparrow$ &
\textbf{Avg. Steps} $\downarrow$ &
\textbf{Tok./Ep.} $\downarrow$ &
\textbf{Tok./Succ.} $\downarrow$ \\

\midrule

\multirow{3}{*}{\textbf{Seen}}
& No Experience
& 68.57
& 18.17
& \textbf{27.11k}
& \underline{40.10k} \\

& MemRL
& \underline{90.00}
& \underline{11.13}
& 114.34k
& 127.04k \\

\rowcolor{gray!10}& \method
& \textbf{96.19}
& \textbf{10.27}
& \underline{38.06k}
& \textbf{39.57k} \\

\midrule

\multirow{3}{*}{\textbf{Unseen}}
& No Experience
& 74.88
& 18.19
& \textbf{27.89k}
& \textbf{37.25k} \\

& MemRL
& \underline{90.30}
& \underline{12.31}
& 125.28k
& 138.73k \\

\rowcolor{gray!10} & \method
& \textbf{90.80}
& \textbf{12.24}
& \underline{48.76k}
& \underline{53.70k} \\

\bottomrule
\end{tabular}
}
\end{wraptable}

Among the experience-based learning baselines, MemRL achieves the strongest overall performance, and we therefore use it as the primary reference for the efficiency analysis. As shown in Tab.~\ref{tab:alfworld_effi}, \method~ improves Avg.@3 from 90.00 to 96.19 on seen tasks and from 90.30 to 90.80 on unseen tasks. These gains are achieved with fewer interaction steps, reducing the average episode length from 11.13 to 10.27 on seen tasks and from 12.31 to 12.24 on unseen tasks. The reduction in token consumption is more substantial. On seen tasks, \method~ reduces Tokens/Episode from 114.34k to 38.06k and Tokens/Success from 127.04k to 39.57k, corresponding to reductions of 66.7\% and 68.9\%, respectively. On unseen tasks, both metrics are reduced by approximately 61\%. Overall, \method~ achieves higher task performance than MemRL while requiring fewer interaction steps and substantially fewer tokens.  We additionally report
the corresponding efficiency results on the other benchmarks in
Appendix~\ref{sec:Per_Eff}.

\subsection{Ablation Studies and Analysis}

\begin{wraptable}[16]{r}{0.60\linewidth}
\centering

\caption{Ablation on ALFWorld seen and unseen tasks.
All results are Avg.@3. Hier. Exp., Util. Ret., and Traj. Ret. denote
Hierarchical Experience, Utility-Aware Experience Retrieval, and Trajectory Retrieval.}
\label{tab:ablation_alfworld}
\scriptsize
\setlength{\tabcolsep}{2.8pt}
\renewcommand{\arraystretch}{0.80}

\resizebox{\linewidth}{!}{
\begin{tabular}{lcccc@{\hspace{4pt}\vrule\hspace{4pt}}cc}

\toprule

\multirow{2}{*}[-0.4ex]{\textbf{Model}}
& \multicolumn{4}{c}{\textbf{Components}}
& \multicolumn{2}{c}{\textbf{ALFWorld}} \\

\cmidrule(lr){2-5}
\cmidrule(lr){6-7}

& \textbf{Hier. Exp.}
& \textbf{Util. Ret.}
& \textbf{TQL}
& \textbf{Traj. Ret.}
& \textbf{Seen}
& \textbf{Unseen} \\

\midrule

\multirow{6}{*}{\textbf{DeepSeek-V4-Flash}}
& \checkmark & \checkmark & \checkmark & \checkmark
& \textbf{96.19} & \textbf{90.80} \\

& \checkmark & \checkmark & \checkmark & \xmark
& 91.43 & 89.80 \\


& \checkmark & \checkmark & \xmark & \xmark
& 78.81 & 79.35 \\

& \checkmark & \xmark & \xmark & \xmark
& 84.76 & 83.08 \\

& \xmark & \checkmark & \checkmark & \checkmark
& 93.57 & 89.55 \\

& \xmark & \xmark & \xmark & \xmark
& 68.57 & 74.88 \\

\addlinespace[1.5pt]
\midrule
\addlinespace[1.5pt]

\multirow{6}{*}{\textbf{Qwen3.5-Flash}}
& \checkmark & \checkmark & \checkmark & \checkmark
& \textbf{93.10} & \textbf{89.05} \\

& \checkmark & \checkmark & \checkmark & \xmark
& 88.10  & 86.32  \\

& \checkmark & \checkmark & \xmark & \xmark
& 76.90 & 82.34 \\

& \checkmark & \xmark & \xmark & \xmark
& 75.95 & 78.11 \\

& \xmark & \checkmark & \checkmark & \checkmark
& 88.60 & 85.57 \\

& \xmark & \xmark & \xmark & \xmark
& 65.00 & 70.40 \\

\bottomrule

\end{tabular}
}

\end{wraptable}

Both Utility-Aware Experience Retrieval and Trajectory Retrieval are built on
the trajectory-quality signal provided by TQL: the former uses TQL to learn
experience utility, while the latter uses TQL to select high-quality successful
trajectories. We therefore conduct a dependency-aware progressive ablation over
these three components to examine their respective contributions. Specifically,
we first remove Trajectory Retrieval, disabling successful trajectory
exemplars. We then replace TQL with the outcome-only signal $1-y$, and finally
remove Utility-Aware Experience Retrieval, reducing experience retrieval to
cosine similarity alone. Hierarchical Experience is ablated separately by
removing task-type-level experiences while retaining task-level experiences.

As shown in Tab.~\ref{tab:ablation_alfworld}, removing Trajectory Retrieval
consistently reduces performance, while replacing TQL with outcome-only
feedback leads to a larger degradation, highlighting the importance of
TQL for experience learning. The effect of
Utility-Aware Experience Retrieval further depends on the quality of its utility signal:
Qwen3.5-Flash degrades further under cosine-only retrieval, whereas
DeepSeek-V4-Flash partially recovers when the outcome-based utility is no
longer used for retrieval. This suggests that, for DeepSeek-V4-Flash, utility
estimated from the coarse $1-y$ signal may introduce noise into
Utility-Aware Experience Retrieval. This difference may stem from the stronger base capability of
DeepSeek-V4-Flash, for which the binary $1-y$ signal becomes less
discriminative because many experiences lead to successful outcomes.
As a result, cosine similarity can be more reliable than retrieval with
coarse utility estimates, while outcome-based utility remains more
informative for Qwen3.5-Flash.
Separately removing task-type-level experience also reduces performance across
both models, showing that reusable local decision patterns complement
task-specific global planning. Overall, the full model achieves the strongest
performance through the combined use of TQL, Utility-Aware Experience Retrieval, Trajectory
Retrieval, and Hierarchical Experience.

\subsection{Effect of relevance-utility trade-off coefficient \texorpdfstring{$\alpha$}{alpha}}

\begin{wrapfigure}[16]{l}{0.49\linewidth}
    \centering
    \vspace{-10pt}
    \includegraphics[
        width=\linewidth,
        height=0.22\textheight,
        keepaspectratio
    ]{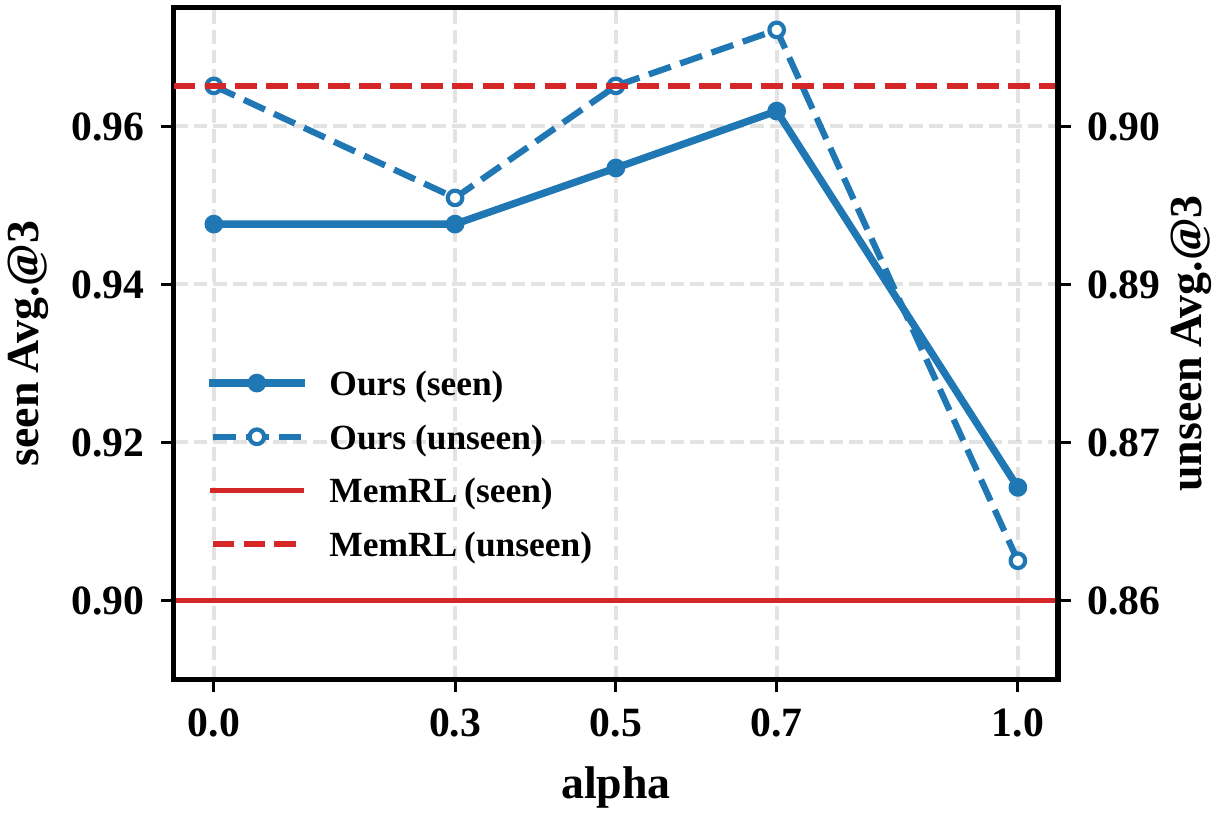}
    \caption{Sensitivity of performance to $\alpha$
    on ALFWorld using DeepSeek-V4-Flash.}
    \label{fig:alpha}
    \vspace{-10pt}
\end{wrapfigure}
Fig.~\ref{fig:alpha} examines the sensitivity of our method
to the relevance-utility trade-off coefficient  $\alpha$ on both seen and unseen
tasks. Performance remains relatively stable for
$\alpha \in [0, 0.7]$, with both splits achieving their
highest performance at $\alpha=0.7$, where our method
outperforms MemRL on both seen and unseen tasks.
However, performance drops noticeably at $\alpha=1$, suggesting that learned
utility provides useful information beyond semantic relevance.
These results indicate that combining semantic relevance
with learned utility improves experience retrieval,
with $\alpha=0.7$ providing the highest observed
performance on both splits. Although performance declines at $\alpha=1$, our method maintains relatively
high performance, particularly on seen tasks, where
it still outperforms MemRL. At $\alpha=1$, TQL-guided experience extraction and trajectory selection, along with task-type-based candidate filtering, remain unchanged, which may help maintain relatively high performance.

\subsection{Epoch-wise Comparison of Trajectory Quality Loss and Task Success Rate}



\begin{wrapfigure}[16]{r}{0.47\linewidth}
    \centering
    \includegraphics[
        width=\linewidth,
        keepaspectratio
    ]{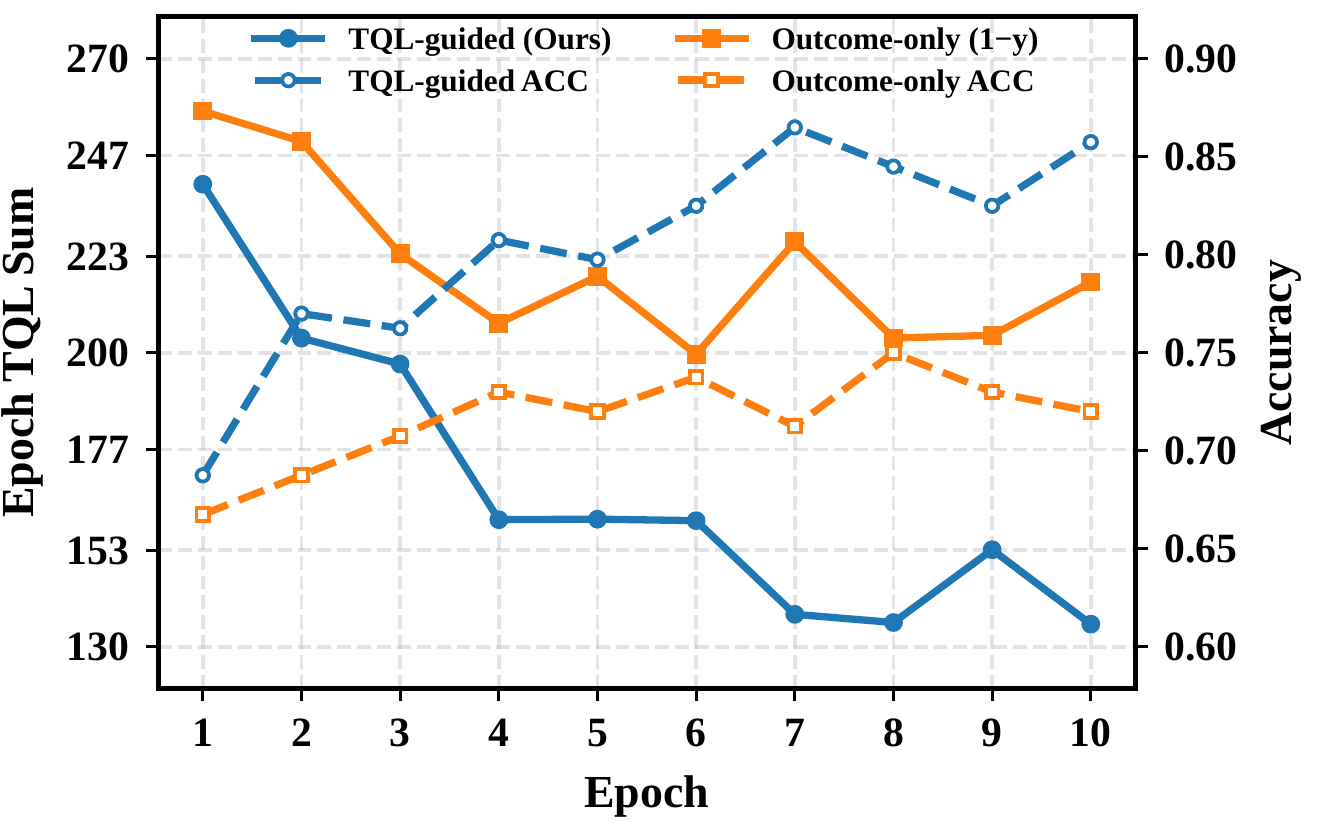}
    \vspace{-20pt}
    \caption{Epoch-wise TQL sum and task success rate
during experience learning. Solid lines indicate
TQL sums (left axis), while dashed lines indicate
task success rates (right axis).}
    \label{fig:epoch-tql-sum}
\end{wrapfigure}

We further examine the evolution of trajectory quality and task
performance during experience learning using the epoch-wise TQL
sum and task success rate. The epoch-wise TQL sum is defined as
the sum of the original trajectory-level TQL values over all
rollouts in the experience-guided groups collected in each epoch,
while the success rate is the average success rate of these rollouts.
To examine the effect of trajectory-quality feedback, we construct
an outcome-only variant in which TQL is replaced by $1-y$,
where $y$ indicates task success. Importantly, $1-y$ is used only
as the learning signal, and both variants are evaluated using
the same original TQL metric to ensure a consistent comparison.

As shown in Fig.~\ref{fig:epoch-tql-sum}, the TQL-guided variant
exhibits a clearer overall decrease in epoch-wise TQL sum,
accompanied by an increasing trend in task success rate.
In contrast, the outcome-only variant shows a smaller improvement
in success rate, while its TQL sum decreases less consistently
and exhibits larger fluctuations in later epochs.
These results suggest that trajectory-quality feedback provides
more informative guidance for experience optimization than
outcome-only feedback, leading to improvements in both overall
trajectory quality and task success.
Additional epoch-wise TQL-sum results on other benchmarks are
provided in Appendix~\ref{sec:additional_tql_results}.

\subsection{Semantic Evolution of Hierarchical Experience}

\begin{figure}[H]
      \centering
      \includegraphics[width=\linewidth]{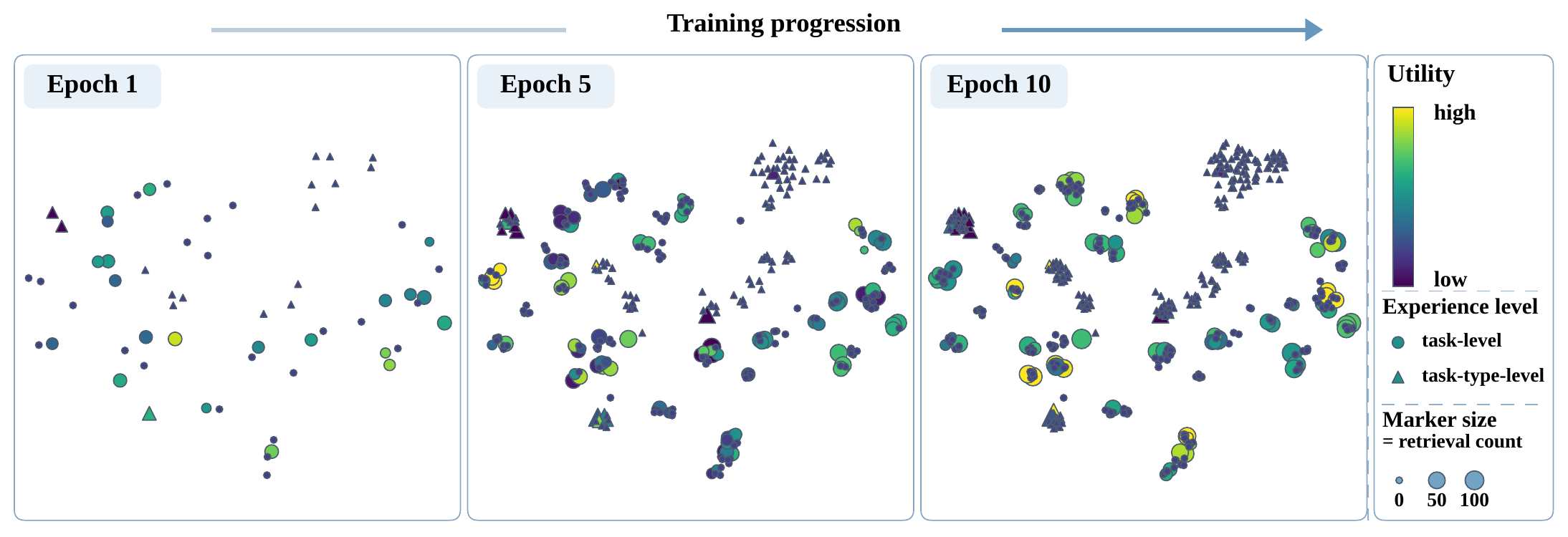}
      \caption{
      Semantic evolution of the ALFWorld experience bank.
      Circles denote task experiences, triangles denote task-type experiences,
      color indicates utility, and marker size indicates retrieval count.
      }
      \label{fig:alfworld_experience_evolution}
\end{figure}

\vspace{-1em}
Fig.\ref{fig:alfworld_experience_evolution} visualizes how the hierarchical
experience space evolves over training on ALFWorld with DeepSeek-V4-Flash as
the backbone LLM. We encode the complete textual content
of each task-level and task-type-level experience using
Qwen3-Embedding-0.6B and jointly project the experience-bank snapshots from
Epochs~1, 5, and 10 into a shared two-dimensional t-SNE space. Utility and raw
retrieval count determine marker color and size, respectively.

The projection reveals clear but distinct clustering structures at the two
experience levels. Task-level experiences form multiple localized clusters
distributed across the semantic space, reflecting diverse task-specific goals
and procedures. Task-type-level experiences also exhibit compact clusters,
but tend to concentrate in more distinct semantic regions, consistent with
their role in capturing reusable stage-level decision patterns. The different
spatial organizations suggest that the two levels encode complementary
knowledge at different levels of abstraction. As training progresses, experiences become increasingly dense
within established semantic regions, indicating that the experience space
gradually shifts from expanding coverage to refining previously acquired
knowledge. Meanwhile, the number of high-utility and frequently retrieved experiences grows, as shown by the increasing prevalence of markers associated with higher utility values and larger marker sizes in later snapshots. Additionally, we analyze the evolution of experience bank 
sizes during training in Appendix 
\ref{sec:Exp_Bank_Evo}.

\section{Conclusion}

In this work, we proposed \method, a trajectory-quality-guided
experience-based learning framework that moves beyond outcome-based
feedback by considering the quality of the execution process.
We introduced Trajectory Quality Loss (TQL) to evaluate both
task completion and process quality, providing a learning
signal for experience optimization.
Using this signal, we employed paired rollouts with and
without retrieved experience to select representative
trajectories for hierarchical experience extraction and
estimate the contribution of retrieved experience.
The estimated utility was then incorporated into
subsequent experience retrieval.
Experiments on ALFWorld, WebShop, and ScienceWorld showed
that \method~consistently outperformed existing
experience-based methods, improving overall Avg.@3
over the strongest baseline by 4.50 and 8.55 percentage
points on DeepSeek-V4-Flash and Qwen3.5-Flash, respectively.
These results highlight the potential of process
feedback for improving experience learning.

\bibliography{iclr2027_conference}
\bibliographystyle{iclr2027_conference}

\appendix

\section{Additional Method Details}

\subsection{Detailed Definition of Trajectory Quality Loss}
\label{sec:tql}

We define Trajectory Quality Loss (TQL) to evaluate the quality of an interaction trajectory by jointly considering task completion and process quality. For a trajectory $\tau_i$ corresponding to task $q$, let $y_i \in \{0,1\}$ indicate whether the task is successfully completed.
The trajectory-level TQL is defined as:
\begin{equation}
\mathcal{L}_{\mathrm{TQL}}(\tau_i)
=
(1-y_i)
+
\left(1-\mathrm{PQ}(\tau_i)\right),
\end{equation}

where $\mathrm{PQ}(\tau_i)\in(0,1]$ denotes the  quality of the execution process. A lower TQL value indicates a higher-quality trajectory.

We measure process quality from four aspects: trajectory length, action validity, repeated-action loops, and the position of the first detected error.  For each trajectory $\tau_i$, let $T_i$ denote the number of interaction steps.
Let $\mu_T$ and $\sigma_T$ denote the mean and standard deviation of the lengths of all trajectories sampled for task $q$.
 We first define the positive length deviation as:
\begin{equation}
z_i^+
=
\max\left(
\frac{T_i-\mu_T}{\sigma_T},
0
\right),
\end{equation}
where $\sigma_T$ is set to $1$ when the empirical standard deviation is zero.   The resulting deviation is then normalized over all trajectories of the same task $q$ to obtain the length penalty:
\begin{equation}
\ell_i =
\begin{cases}
\dfrac{z_i^+}{\max_j z_j^+}, & \text{if } \max_j z_j^+ > 0, \\
0, & \text{otherwise}.
\end{cases}
\end{equation}

We next measure action validity. Let $v_{i,t}\in\{0,1\}$ indicate whether the action at step $t$ can be parsed and accepted as a valid executable action by the corresponding environment. The action-validity ratio is defined as:

\begin{equation}
  r_i^{\mathrm{valid}}
  =
  \frac{1}{T_i}
  \sum_{t=1}^{T_i}
  v_{i,t}.
\end{equation}


To account for repetitive behavior, we further introduce a repeated-action loop penalty. We define a repeated-action loop as an immediately repeated contiguous block of actions. Specifically, for a period $p \in \{1,\ldots,P\}$ and a start  position $s$, the action subsequence
$(a_{i,s}, a_{i,s+1}, \ldots, a_{i,s+p-1})$
is considered to form a repeated-action loop if it is immediately followed by an identical subsequence:
\begin{equation}
  (a_{i,s}, a_{i,s+1}, \ldots, a_{i,s+p-1})
  =
  (a_{i,s+p}, a_{i,s+p+1}, \ldots, a_{i,s+2p-1}).
\end{equation}
Here, $p$ denotes the loop period, and $P$ is the maximum period considered. For example, period-$1$ loops capture repetitions such as ``open fridge, open fridge'', while period-$2$ loops capture alternating patterns such as ``go north, go south, go north, go south''. When such a match is found, we mark only the later copy of the repeated block,
$\{s+p,\ldots,s+2p-1\}$, as loop positions. Since loop positions are represented as a set over all detected matches across different starts $s$ and periods $p \le P$, an action position marked multiple times contributes only once to
$n_i^{\mathrm{loop}}$.

Let $n_i^{\mathrm{loop}}$ denote the number of unique action positions in trajectory $\tau_i$ that are marked as belonging to such immediately repeated blocks. The loop penalty is defined as:
\begin{equation}
  r_i^{\mathrm{loop}}
  =
  \frac{n_i^{\mathrm{loop}}}{\max(T_i-1,1)}.
\end{equation}

We also consider when the first error occurs. Let $t_i^{\mathrm{err}}$ denote the first step at which an error is detected. A step is regarded as erroneous if it contains an invalid action,  an observation indicating that nothing happens, or an action belonging to a repeated-action loop. The early-error penalty is defined as:
\begin{equation}
p_i^{\mathrm{err}}
=
\begin{cases}
0, & \text{if no error is detected}, \\
1-\dfrac{t_i^{\mathrm{err}}}{T_i}, & \text{otherwise}.
\end{cases}
\end{equation}
This assigns a larger penalty to errors that occur earlier in the trajectory. The process quality score is then computed as:
\begin{equation}
\mathrm{PQ}(\tau_i)
=
\exp\left(
- \ell_i
-\left(1-r_i^{\mathrm{valid}}\right)
- r_i^{\mathrm{loop}}
- p_i^{\mathrm{err}}
\right),
\end{equation}
These four components capture complementary and environment-agnostic aspects of interactive process quality: efficiency (trajectory length), executability (action validity), behavioral stability(repeated-action loops), and progress before failure (first-error position). Importantly, they are computed directly from environment interactions without requiring an additional learned rewardmodel or task-specific process annotations.

\subsection{Experience Extraction}
\label{sec:exp_extra}
\subsubsection{Task-Level Experience Extraction}
\label{sec:task_exp}

For each task $q$, the paired rollouts produce an experience-guided group
$\mathcal{G}^{\mathrm{exp}}$ and a baseline group
$\mathcal{G}^{\mathrm{base}}$. Reflecting on all sampled trajectories is often
unnecessary and may introduce redundant evidence. We therefore select a small
set of representative trajectories that reveal either effective behavior or
informative mistakes. Let:
\begin{equation}
\mathcal{G}^{all}
=
\mathcal{G}^{\mathrm{exp}}
\cup
\mathcal{G}^{\mathrm{base}}
\end{equation}
denote all sampled trajectories for the task. Each trajectory $\tau_i$ has a
success indicator $y_i$ and a trajectory-level TQL value
$\mathcal{L}_{\mathrm{TQL}}(\tau_i)$, where lower TQL indicates a better trajectory.

We first collect all successful trajectories:
\begin{equation}
\mathcal{G}^{+}
=
\left\{
\tau_i \in \mathcal{G}^{all}
:
y_i = 1
\right\}.
\end{equation}
We also divide the baseline trajectories into successful and failed subsets:
\begin{equation}
\mathcal{G}^{\mathrm{base},+}
=
\left\{
\tau_i \in \mathcal{G}^{\mathrm{base}}
:
y_i = 1
\right\},
\qquad
\mathcal{G}^{\mathrm{base},-}
=
\left\{
\tau_i \in \mathcal{G}^{\mathrm{base}}
:
y_i = 0
\right\}.
\end{equation}
We draw failure representatives from the baseline group because these failures
show what the agent cannot reliably handle without experience support. Distilling such failures helps the experience capture mistakes that future retrieved
experiences should prevent. In contrast, the best successful representatives may come from either group, since both baseline and experience-guided successes can provide evidence of effective behavior.

When the corresponding sets are non-empty, we choose representatives according
to success status and TQL. Among all successful trajectories, we select the
cleanest successful solution:
\begin{equation}
\tau^{\mathrm{best}}_{\mathrm{succ}}
=
\arg\min_{\tau_i \in \mathcal{G}^{+}}
\mathcal{L}_{\mathrm{TQL}}(\tau_i).
\end{equation}
Among failed baseline trajectories, we select two failures with different
degrees of quality:
\begin{equation}
\tau^{\mathrm{near}}_{\mathrm{fail}}
=
\arg\min_{\tau_i \in \mathcal{G}^{\mathrm{base},-}}
\mathcal{L}_{\mathrm{TQL}}(\tau_i),
\qquad
\tau^{\mathrm{bad}}_{\mathrm{fail}}
=
\arg\max_{\tau_i \in \mathcal{G}^{\mathrm{base},-}}
\mathcal{L}_{\mathrm{TQL}}(\tau_i).
\end{equation}
The first is the least severe failed attempt and may contain partial progress,
whereas the second captures a more severe failure pattern. If baseline
trajectories all succeed, failures are unavailable; in this case, we select the
least efficient successful baseline trajectory:
\begin{equation}
\tau^{\mathrm{worst}}_{\mathrm{succ}}
=
\arg\max_{\tau_i \in \mathcal{G}^{\mathrm{base},+}}
\mathcal{L}_{TQL}(\tau_i).
\end{equation}

These representatives are used according to the rollout outcome. For  task $q$, we apply an ordered decision rule to its paired rollout group
$\mathcal{G}^{all}=\mathcal{G}^{\mathrm{exp}}\cup\mathcal{G}^{\mathrm{base}}$.
The rule selects at most one reflection mode. If $\mathcal{G}^{all}$ contains at
least one successful trajectory and the baseline group
$\mathcal{G}^{\mathrm{base}}$ contains at least one failed trajectory, we use
failure-aware reflection. If no trajectory in $\mathcal{G}^{all}$ succeeds but
$\mathcal{G}^{\mathrm{base}}$ contains failed trajectories, we use failure-only
reflection. If all available baseline trajectories succeed, we consider
success-contrast reflection only when the baseline group-level TQL exceeds a
threshold. In all remaining cases, we skip reflection because the sampled
 rollouts provide insufficient new evidence for experience extraction.

\textbf{Case 1: Failure-aware reflection.}
This case is selected when $\mathcal{G}^{+} \neq \emptyset$ and $\mathcal{G}^{\mathrm{base},-} \neq \emptyset$, where $\mathcal{G}^{+}$ contains successful trajectories from either the experience-guided or baseline group. We compare the best successful trajectory $\tau^{\mathrm{best}}_{\mathrm{succ}}$ with the near-miss and severe baseline failures, $\tau^{\mathrm{near}}_{\mathrm{fail}}$ and $\tau^{\mathrm{bad}}_{\mathrm{fail}}$. The successful trajectory reveals effective task-solving steps, the near-miss failure can help uncover overlooked checks or incomplete actions, and the severe failure can expose serious error patterns to avoid.

\textbf{Case 2: Failure-only reflection.}
This case is selected when $\mathcal{G}^{+} = \emptyset$ and  $\mathcal{G}^{\mathrm{base},-} \neq \emptyset$. We compare $\tau^{\mathrm{near}}_{\mathrm{fail}}$ and  $\tau^{\mathrm{bad}}_{\mathrm{fail}}$. Although no successful solution is
 observed, this comparison can still reveal partial progress, missing checks,
and recurring error patterns by contrasting a less severe failure with a more
severe one.

\textbf{Case 3: Success-contrast reflection.}
This case is considered when
  $\mathcal{G}^{\mathrm{base},-} = \emptyset$ and
  $\mathcal{G}^{\mathrm{base},+} \neq \emptyset$.
  Because all baseline trajectories succeed, reflection is performed only when
  their behavior remains inefficient or unstable. We quantify this using the
  baseline group-level TQL:
\begin{equation}
\mathcal{L}_{\mathrm{TQL}}(\mathcal{G}^{base})
=
\frac{1}{\left|\mathcal{G}^{\mathrm{base}}\right|}
\sum_{\tau_i \in \mathcal{G}^{\mathrm{base}}}
\mathcal{L}_{TQL}(\tau_i)
+
\left(
1
-
\frac{1}{\left|\mathcal{G}^{\mathrm{base}}\right|}
\sum_{\tau_i \in \mathcal{G}^{\mathrm{base}}}
y_i
\right).
\end{equation}

If  $\mathcal{L}_{\mathrm{TQL}}(\mathcal{G}^{base})\geq 0.7$, we compare $\tau^{\mathrm{best}}_{\mathrm{succ}}$ with
$\tau^{\mathrm{worst}}_{\mathrm{succ}}$ to extract lessons about efficient
action ordering, avoiding redundant actions, and stabilizing successful
behavior. Otherwise, we skip reflection because the baseline successes already
provide little new information.

For each task $q$, the selected representatives, when available, are passed to  the reflection model, yielding at most one reflection request for the paired rollout group  $\mathcal{G}=\mathcal{G}^{\mathrm{exp}}\cup\mathcal{G}^{\mathrm{base}}$. The model distills the representatives into one to three ordered task-stage experiences. Each stage specifies when the lesson should be applied, the principle to follow, and, when applicable, the failure pattern it is intended to prevent. The resulting task-level experience is stored in $\mathcal{E}^{\mathrm{task}}$ and buffered according to task type for later abstraction into task-type-level experience.

\subsubsection{Task-Type Experience Distillation}
\label{sec: task-type-exp}

Task-level experiences summarize the solution patterns observed in individual task instances. However, they may contain instance-specific details that limit their transferability. Since tasks of the same type often share common stages, decision patterns, and failure modes, we periodically distill task-level experiences into reusable task-type experiences.

Let $\phi(q)$ denote the task type of task $q$. For each task type $z$, we
maintain a temporary buffer
\begin{equation}
\mathcal{B}^{\mathrm{task}}_z
=
\left(
\mathcal{E}_z,
\mathcal{Q}_z
\right),
\end{equation}
where $\mathcal{E}_z$ contains valid task-level experiences extracted from tasks
satisfying $\phi(q)=z$, and $\mathcal{Q}_z$ contains their corresponding
task-level TQL values. An experience is added to the buffer only when
task-level reflection produces a valid result.

The buffer is processed once it contains enough distinct task instances or
accumulates sufficient TQL evidence:
\begin{equation}
\left|\mathrm{UniqueTasks}(\mathcal{E}_z)\right|
\geq M_{\min}
\quad\lor\quad
\sum_{q\in\mathrm{UniqueTasks}(\mathcal{E}_z)}
\mathrm{TQL}_q
>
\delta_{\mathrm{type}}.
\end{equation}
After processing, the buffer is cleared and a new buffer is created for later
tasks. In our implementation, $M_{\min}$ and $\delta_{\mathrm{type}}$ are
configurable and are set to $4$ and $5.0$ by default.


Given the buffered staged task-level experiences and their associated TQL values, a reflection model extracts common and reusable patterns across tasks of the same type. The model operates only on the provided task-level experiences and does not directly access the raw trajectories. It merges
semantically similar stages even when their wording, objects, products, or environment details differ, while avoiding the reproduction of a single concrete task unless it expresses a reusable task-type pattern.



For each processed buffer, the reflection model generates up to $K_{\max}$
candidate experiences:
\begin{equation}
\mathcal{C}_z
=
\left\{
e^{\mathrm{type}}_{z,1},
\dots,
e^{\mathrm{type}}_{z,K_z}
\right\},
\qquad
1 \leq K_z \leq K_{\max},
\end{equation}
where $K_{\max}=3$ by default. Candidates with duplicate content
 are removed before being inserted into memory.

Each remaining candidate is then compared with the existing task-type experiences of type $z$. Based on the candidate and the existing memory, the same model used for trajectory execution selects one of four operations: add, merge, modify, or discard.

\textbf{Add.}
A candidate is added when it introduces a new task subtype, stage, ordering
pattern, critical step, or failure mode that is not covered by the existing
memory:
\begin{equation}
\mathcal{E}^{\mathrm{type}}_z
\leftarrow
\mathcal{E}^{\mathrm{type}}_z
\cup
\left\{
e^{\mathrm{type}}
\right\}.
\end{equation}
If no task-type experience of type $z$ exists, the candidate is added directly.

\textbf{Merge.}
A candidate is merged with an existing experience when both express the same
main principle but the candidate provides additional useful details, such as
new critical steps, failure patterns, or avoidance rules:
\begin{equation}
e_j^{\mathrm{type}}
\leftarrow
\mathrm{Merge}
\left(
e_j^{\mathrm{type}},
e^{\mathrm{type}}
\right).
\end{equation}

\textbf{Modify.}
A candidate is used to revise an existing experience when the current
experience is misleading, overly broad, overly narrow, outdated, or
inconsistent with the buffered evidence:
\begin{equation}
e_j^{\mathrm{type}}
\leftarrow
\mathrm{Modify}
\left(
e_j^{\mathrm{type}},
e^{\mathrm{type}}
\right).
\end{equation}
The revised experience retains the same memory identifier, increments its
version, and decays its utility to reflect the uncertainty introduced by the
update.

\textbf{Discard.}
A candidate is discarded when it is redundant, too generic, or does not provide
actionable information beyond the existing memory:
\begin{equation}
\mathcal{E}^{\mathrm{type}}_z
\leftarrow
\mathcal{E}^{\mathrm{type}}_z.
\end{equation}

\subsection{Trajectory Collection and Retrieval}
\label{sec:trajectory_retrieval}

We construct a trajectory pool from the training tasks for use as execution
exemplars during evaluation. The procedure starts from trajectories retained
at the final training epoch, followed by an additional collection stage and
trajectory update.

\textbf{Initial Trajectory Selection.}
For each training task $q$, we retain the successful trajectory with the
minimum TQL from the final training epoch:
\begin{equation}
\label{eq:initial_retained_traj}
\tau_q^{\mathrm{ret}}
=
\arg\min_{\tau\in\mathcal{G}_q^{\mathrm{final}},\,y(\tau)=1}
\mathcal{L}_{\mathrm{TQL}}(\tau),
\end{equation}
where $\mathcal{G}_q^{\mathrm{final}}$ denotes the trajectories generated for
$q$ at the final training epoch. These trajectories form the initial trajectory pool.

\textbf{Trajectory Collection.}
We then perform an additional collection stage over the training tasks.
For each task $q$, the agent retrieves learned experiences together with
previously retained successful trajectories from similar tasks and uses them
as execution context. Among the successful trajectories generated in this
stage, the one with the minimum TQL is selected as the current candidate:
\begin{equation}
\label{eq:candidate_traj}
\tau_q^{\mathrm{cand}}
=
\arg\min_{\tau\in\mathcal{G}_q^{\mathrm{col}},\,y(\tau)=1}
\mathcal{L}_{\mathrm{TQL}}(\tau),
\end{equation}
where $\mathcal{G}_q^{\mathrm{col}}$ denotes the trajectories generated during
trajectory collection.

\textbf{Trajectory Update.}
For each training task $q$, the current candidate
$\tau_q^{\mathrm{cand}}$ is compared with the previously retained trajectory
$\tau_q^{\mathrm{ret}}$. Since both are successful, we compare them
lexicographically by episode length, invalid actions, and corrected actions.
\begin{equation}
\label{eq:trajectory_update}
\tau_q^{*}
=
\operatorname{Better}
\left(
\tau_q^{\mathrm{ret}},
\tau_q^{\mathrm{cand}}
\right).
\end{equation}
Applying this procedure to all training tasks yields the final trajectory pool.

\textbf{Trajectory Retrieval.}
During evaluation, we rank training tasks by semantic similarity to the current
task and retrieve successful trajectories from the top-$K_{\mathrm{tra}}$ most
similar tasks as execution exemplars. Trajectory retrieval is based only on
task similarity and is separate from the Utility-Aware Experience Retrieval
described in Section~\ref{sec:Traj_Qua_Gui}.

\section{Experiment Details}

\subsection{Dataset Composition}
\label{sec:data_setting}

We evaluate \method~on three interactive benchmarks for LLM agents: ALFWorld \citep{shridhar2020alfworld}, which consists of text-based household tasks requiring agents to navigate environments and manipulate objects through textual commands; WebShop \citep{yao2022webshop}, which evaluates agents on online shopping tasks involving product search and purchase decisions under user-specified requirements; and ScienceWorld \citep{wang2022scienceworld}, which contains interactive science tasks that require agents to navigate rooms, manipulate objects, and conduct experiments based on scientific commonsense.

For each benchmark, we construct training and validation sets
containing 100 and 30 instances, respectively.
For ALFWorld, these instances are sampled according to the
original data distribution, while evaluation is conducted
on the original Seen and Unseen test sets. For WebShop and ScienceWorld, sampling from the original
data distribution would leave many task categories
underrepresented or entirely absent. Following the category-based evaluation settings in
\citet{gao2025efficient, yang2025coarse},
we select the 20 largest task categories in WebShop
and the 10 largest in ScienceWorld.
For each benchmark, we sample the training and validation
instances, along with 100 test instances,
from the selected categories.

\subsection{Baseline Summaries}
\label{sec:baseline_sum}

\paragraph{Baselines.} We consider five representative baselines:
\begin{itemize}
    \item \textbf{No Experience.}
    The agent solves each task independently without using external experience from previous tasks. It therefore reflects the performance of the backbone LLM without experience augmentation.

    \item \textbf{ReasoningBank} \citep{ouyang2025reasoningbank}.
    ReasoningBank distills past trajectories into structured memory items that capture useful reasoning strategies and lessons from failures. For each new task, it retrieves the top-$k$ most relevant memories using embedding similarity and adds them to the agent's system instruction. After the task is completed, an LLM-as-a-judge estimates whether the trajectory succeeds or fails, and the trajectory is then used to extract new memory items for future tasks.

    \item \textbf{MemP} \citep{fang2025memp}.
    MemP extracts procedural memory from the trajectories and rewards of previously completed tasks. For a new task, it retrieves the memory associated with the most similar previous task using cosine similarity over task embeddings and provides it to the agent during execution. The memory library is maintained according to task feedback through addition, deletion, and modification of memory entries.

    \item \textbf{MemRL} \citep{zhang2026memrl}.
    MemRL maintains an Intent-Experience-Utility memory, where each entry stores past task experience, including the complete solution trace or execution trajectory, together with a learned utility score. For a new task, semantically similar memories are first retrieved by embedding similarity and then reranked using both similarity and utility. After execution, the utilities of the retrieved memories are updated according to environmental rewards, and new experience entries are constructed from the completed trajectories and added to the memory bank.

    \item \textbf{DeltaMem} \citep{tan2026deltamem}.
    DeltaMem maintains two separate memory structures: a Task-Tree for procedural skills and an Env-Tree for environment knowledge. Each episode is distilled into structured experience, and when a similar memory already exists, only the residual difference is stored instead of a complete trajectory. During retrieval, nodes are matched by embedding similarity with an additional penalty for failed experiences, and the root-to-node path is reconstructed as the context for the agent. Frequently successful residual paths are further consolidated into new root memories.
\end{itemize}

For all baselines, we follow the hyperparameter settings reported in
their original papers. Our method uses eight rollouts per task, evenly
divided between the baseline and experience-guided groups. We set
$\alpha=0.7$ and $c=0.05$, retrieving the top-3 task-level and top-1
task-type-level experiences. For trajectory retrieval, we use the
top-1 trajectory on WebShop and the top-2 on the other datasets.
All methods are trained for 10 epochs, with a maximum of 30 interaction steps per task across the three datasets.

\subsection{Performance and Interaction Efficiency across benchmarks}
\label{sec:Per_Eff}

\begin{table}[H]   
\centering   
\caption{Task Performance and Interaction Efficiency across Benchmarks using DeepSeek-V4-Flash.}    
\label{tab:across_bench_eff}   

\scriptsize
\setlength{\tabcolsep}{2.5pt}   
\renewcommand{\arraystretch}{0.92}   

\resizebox{0.75\linewidth}{!}{   
\begin{tabular}{llcccc}   
\toprule   
\multirow{2}{*}{\textbf{Split}} &   
\multirow{2}{*}{\textbf{Method}} &   
\multirow{2}{*}{\textbf{Average@3} $\uparrow$} &   
\multirow{2}{*}{\textbf{Avg. Steps} $\downarrow$} &   
\multicolumn{2}{c}{\textbf{Token Efficiency} } \\   
\cmidrule(lr){5-6}   
& & & &   
\textbf{Tokens / Episode}  $\downarrow$&   
\textbf{Tokens / Success} $\downarrow$\\   
\midrule   

\multirow{3}{*}{\textbf{Webshop}}   
& No Experience & 42.33 & 21.59 &\textbf{47.20k} 
&\textbf{111.49k}  \\   
& MemRL & \underline{73.67} & \underline{12.60} & 205.00k & 278.28k \\   
\rowcolor{gray!10} & \method & \textbf{79.67} & \textbf{8.77} & \underline{55.90k} & \underline{70.16k} \\   
\midrule   

\multirow{3}{*}{\textbf{Scienceworld}}   
& No Experience & 47.00 & 18.77 & \textbf{30.46k} & \textbf{64.80k} \\   
& MemRL & \underline{69.33} & \underline{14.41} & 220.61k & 318.18k \\   
\rowcolor{gray!10} & \method & \textbf{74.67} & \textbf{12.46} & \underline{60.82k} & \underline{81.45k} \\   
\bottomrule   
\end{tabular}   
}   
\end{table}

We further compare the task performance and interaction
efficiency of \method~ with MemRL on WebShop and ScienceWorld,
as shown in Tab.~\ref{tab:across_bench_eff}. On WebShop, \method~ improves Average@3 from 73.67 to 79.67
while reducing the average episode length from 12.60 to 8.77.
Meanwhile, Tokens/Episode and Tokens/Success decrease by
72.7\% and 74.8\%, respectively. Similar improvements are observed on ScienceWorld, where
\method~ increases Average@3 from 69.33 to 74.67 and reduces
the average episode length from 14.41 to 12.46. Tokens/Episode and Tokens/Success are further reduced
by 72.4\% and 74.4\%, respectively. These results demonstrate that \method~ achieves higher
task performance than MemRL while requiring fewer
interaction steps and substantially fewer tokens
across both benchmarks.

\subsection{TQL-Sum Evolution on Additional Benchmarks}
\label{sec:additional_tql_results}

\begin{figure}[h]
      \centering
      \begin{subfigure}[t]{0.47\textwidth}
          \centering
          \includegraphics[width=\linewidth]{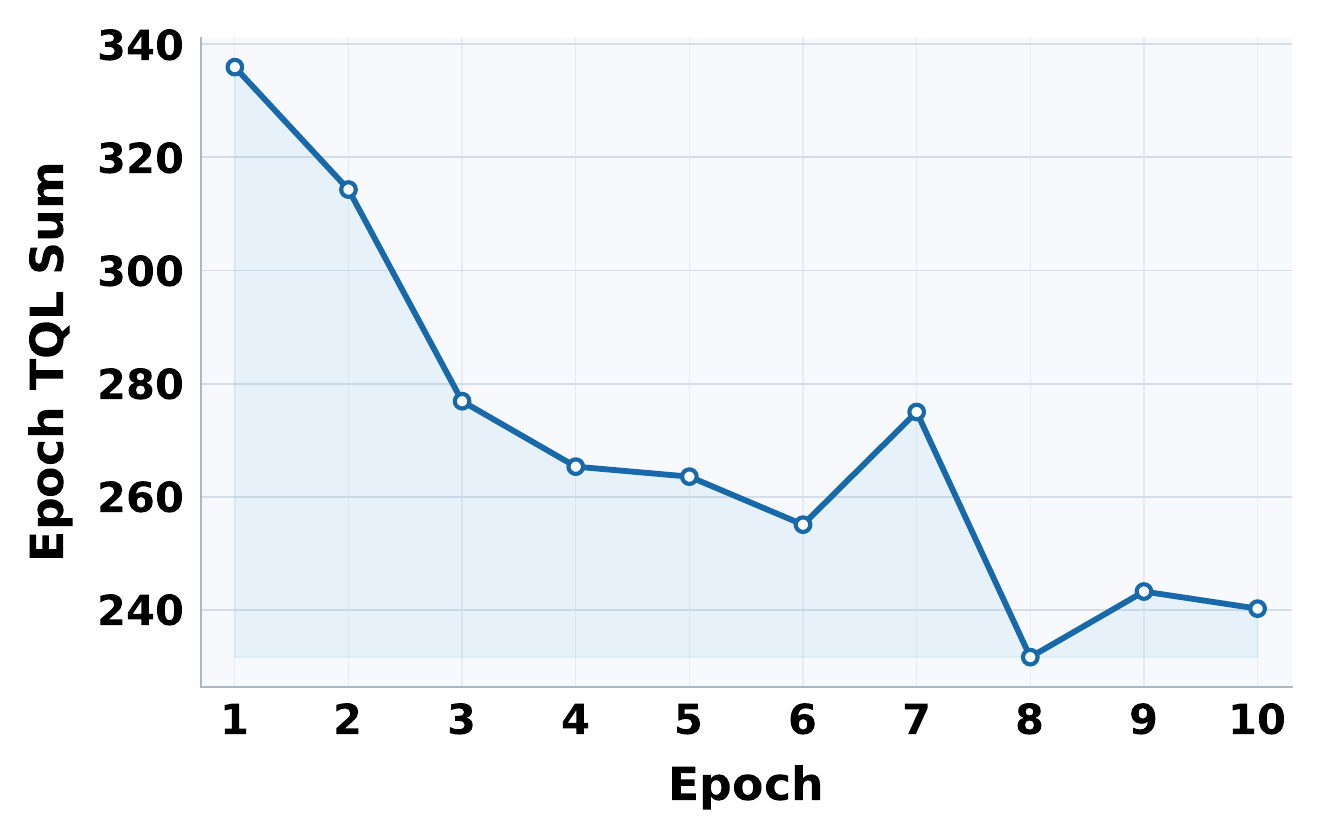}
          \caption{WebShop}
          \label{fig:webshop-tql}
      \end{subfigure}
      \hfill
      \begin{subfigure}[t]{0.47\textwidth}
          \centering
          \includegraphics[width=\linewidth]{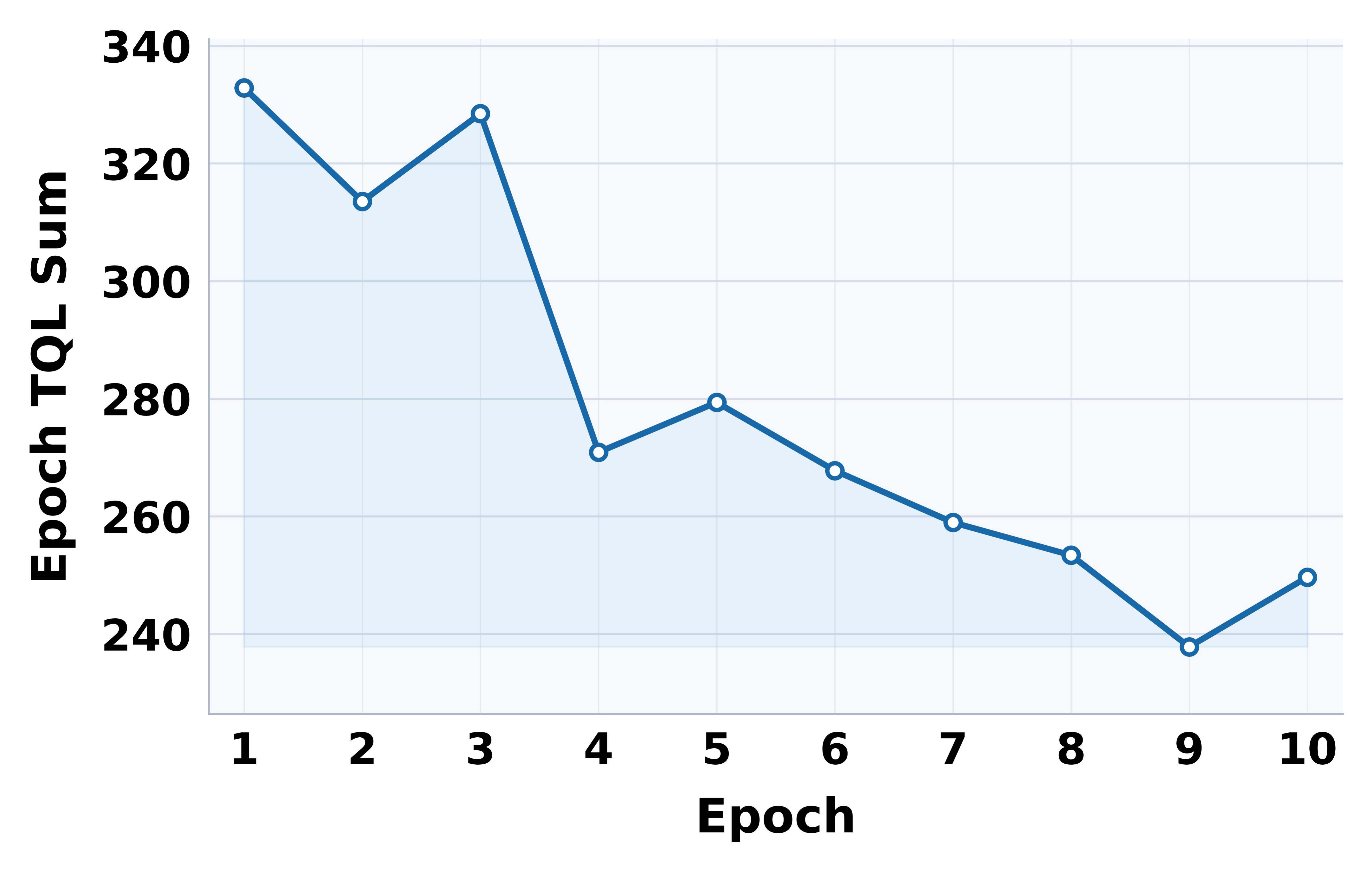}
          \caption{ScienceWorld}
          \label{fig:scienceworld-tql}
      \end{subfigure}

      \caption{Evolution of the epoch-level TQL sum on WebShop and ScienceWorld during training with DeepSeek-V4-Flash.}
      \label{fig:epoch-tql-sum-addition-benchmark}
\end{figure}

Fig.~\ref{fig:epoch-tql-sum-addition-benchmark} shows the evolution of the
epoch-level TQL sum of the experience-guided groups during training on
WebShop and ScienceWorld with DeepSeek-V4-Flash. Across both benchmarks,
the TQL sum follows an overall downward trend, with moderate fluctuations
across epochs. This behavior resembles the optimization of a training
objective: although the signal does not decrease monotonically, its overall
reduction indicates that the utility updates increasingly favor experiences
associated with higher-quality trajectories.

\subsection{Experience-Bank Evolution}
\label{sec:Exp_Bank_Evo}

\begin{wrapfigure}[15]{r}{0.47\textwidth}
    \centering
    \vspace{-14pt}
    \includegraphics[width=\linewidth]{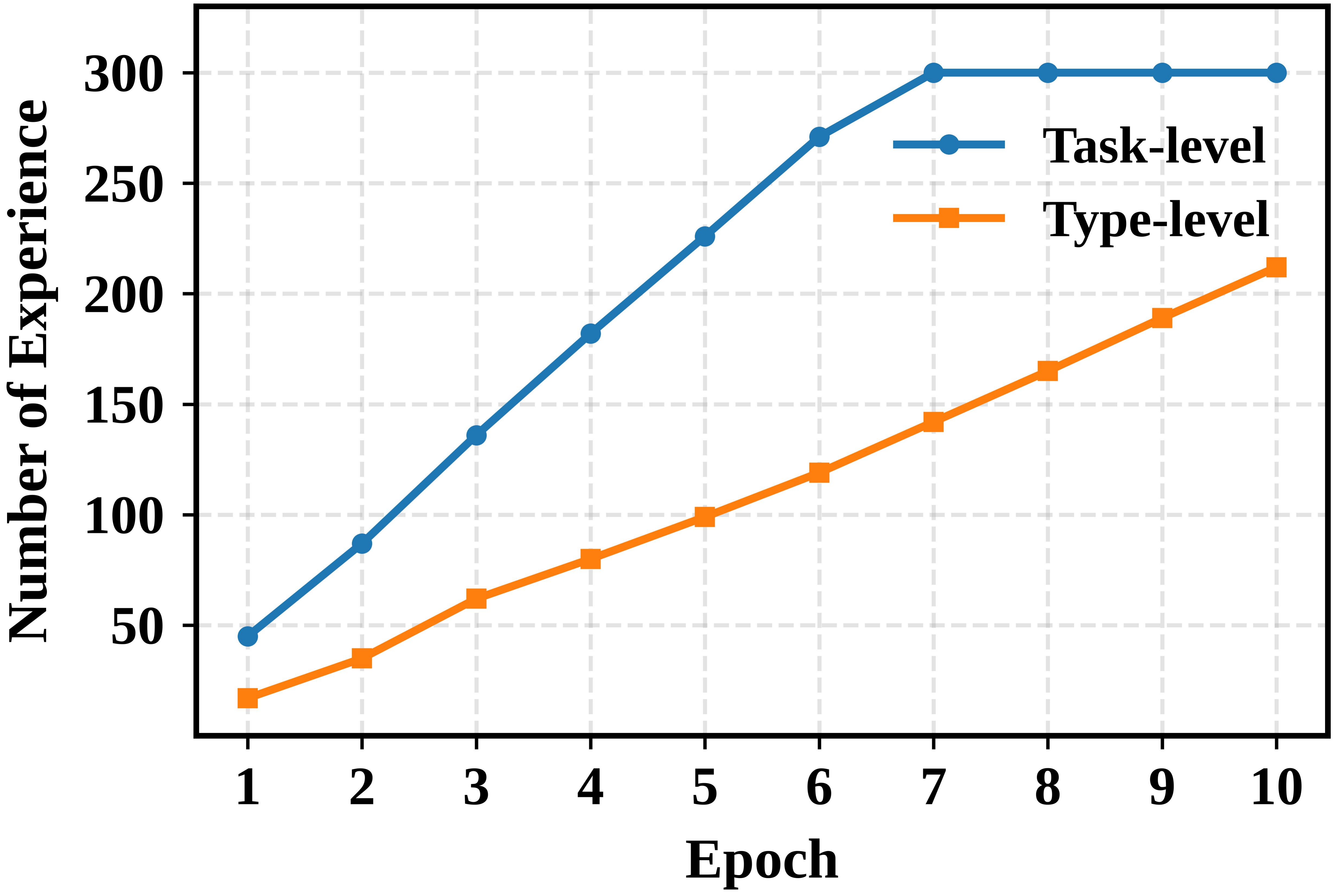}
    \vspace{-1em}
    \caption{Sizes of the task-level and task-type-level experience banks on ALFWorld using DeepSeek-V4-Flash.}
    \label{fig:experience-bank-evolution}
    \vspace{-8pt}
\end{wrapfigure}

Fig.~\ref{fig:experience-bank-evolution} shows the evolution of the task-level and task-type-level experience bank sizes during training on ALFWorld with DeepSeek-V4-Flash. The task-level bank increases from 45 experiences at 
Epoch~1 to its capacity of 300 at Epoch~7, after which its size remains fixed. 
In contrast, the type-level bank grows more gradually, from 17 experiences at 
Epoch~1 to 212 experiences at Epoch~10, without reaching the task-level capacity. Once the task-level experience bank reaches capacity,
newly generated experiences are evaluated alongside
existing ones using the UCB-based score
$u(e) + c\sqrt{\frac{\log(1+N)}{1+n(e)}}$.
Experiences with the lowest scores are removed first. Consequently, from Epoch~7 
onward, the task-level bank maintains a nearly constant size of 300, while 
lower-ranked experiences are gradually replaced by newly generated ones.

\section{Experience Extraction Prompts and Examples}
\label{sec:prompt_and_Exam}

\subsection{Task-Level Experience Extraction Prompt}
\label{sec:task_level_prompt}

\begin{promptbox}{Failure-Aware Reflection}

\textbf{Role.}

You are summarizing ordered stages for one task-level experience from representative trajectories.

\medskip
\textbf{Task Information.}

\begin{quote}
\ttfamily
Task: \{item.get("task", "")\}

Task Type: \{item.get("task\_type", "")\}

Environment/Dataset: \{env\_name\}

Missing representative types: \{json.dumps(missing, ensure\_ascii=False)\}
\end{quote}

\medskip
\textbf{Representative Trajectories.}

\begin{quote}
\ttfamily
\{self.\_format\_optional\_representative(
"BEST\_SUCCESS\_TRAJECTORY",
item.get("tau\_best\_success"))\}

\{self.\_format\_optional\_representative(
"NEAR\_MISS\_FAILURE\_TRAJECTORY",
item.get("tau\_near\_failure"))\}

\{self.\_format\_optional\_representative(
"SEVERE\_FAILURE\_TRAJECTORY",
item.get("tau\_bad\_failure"))\}
\end{quote}

\medskip
\textbf{Reflection Instruction.}

Failure representatives are selected from baseline trajectories without retrieved experience, so use them to diagnose the base model's task mistakes.

Produce 1 to 3 ordered TASK\_STAGE items as a JSON array.

Each item must describe one distinct, reusable stage-level lesson for this concrete task.
Prefer causal mistakes over later symptoms: summarize a later invalid, repeated, or final wrong action only when it reveals an earlier missing check, wrong decision, or incorrect task understanding.
Order items by when the lesson should be applied during task execution.

\medskip
\textbf{Required JSON Fields.}

\begin{itemize}
    \item \texttt{"order"}: integer, starting from 1
    \item \texttt{"stage"}: short environment-appropriate stage name
    \item \texttt{"title"}: short name of the experience
    \item \texttt{"principle"}: the concrete lesson
    \item \texttt{"when\_to\_apply"}: when this lesson should be used
\end{itemize}

\textbf{Optional Fields.}

\begin{itemize}
    \item \texttt{"evidence"}: \texttt{"best\_success"} $\mid$
    \texttt{"near\_miss"} $\mid$
    \texttt{"severe\_failure"} $\mid$
    \texttt{"mixed"}
    \item \texttt{"failure\_to\_avoid"}: common mistake this lesson prevents
\end{itemize}

\textbf{Rules.}

\begin{itemize}
    \item Summarize only stage-level lessons for this concrete task.
    \item Do not force any fixed stage schema. Use stage names appropriate to the environment and task.
    \item Preserve execution order when the trajectory shows a clear order.
    \item Identify the earliest critical mistake or divergence that made later failure likely, and summarize the correction at the appropriate task stage.
    \item Use successful trajectories to identify effective key steps.
    \item Use near-miss failures to identify missing critical actions.
    \item Use severe failures to identify error patterns to avoid.
    \item The experience should be reusable for the same concrete task, not a verbatim copy of actions.
    \item Return only a valid JSON array and no extra text.
\end{itemize}

\end{promptbox}

\begin{promptbox}{Failure-Only Reflection}

\textbf{Role.}

You are summarizing ordered avoidance lessons for one task-level experience from failed trajectories only.

\medskip
\textbf{Task Information.}

\begin{quote}
\ttfamily
Task: \{item.get("task", "")\}

Task Type: \{item.get("task\_type", "")\}

Environment/Dataset: \{env\_name\}

Selection case: failure\_only

Missing representative types: \{json.dumps(missing, ensure\_ascii=False)\}
\end{quote}

\medskip
\textbf{Representative Trajectories.}

\begin{quote}
\ttfamily
\{self.\_format\_optional\_representative(
"LOWEST\_TQL\_FAILURE\_TRAJECTORY",
item.get("tau\_near\_failure"))\}

\{self.\_format\_optional\_representative(
"HIGHEST\_TQL\_FAILURE\_TRAJECTORY",
item.get("tau\_bad\_failure"))\}
\end{quote}

\medskip
\textbf{Reflection Instruction.}

The failed trajectories are selected from baseline attempts without retrieved experience. Treat them as evidence of base-model task mistakes.

Compare the failed trajectories for the same task.

The lowest-TQL failure is the least bad failed attempt and may show partial progress.

The highest-TQL failure is the worst failed attempt and may show severe error patterns.

Extract reusable task-level experience about what to avoid and how to correct the next attempt.

Produce 1 to 3 ordered TASK\_STAGE items as a JSON array.

Prefer early task stages where the failed trajectories begin to drift toward failure.

These items will be stored together as stages inside one task-level experience.

Each item must describe one distinct, reusable avoidance or correction lesson for this concrete task.

Prefer earlier causal mistakes when the evidence supports them, but do not force an early cause if the failure evidence is only visible later.

The items must be ordered by when the correction should be applied during task execution.

\medskip
\textbf{Required JSON Fields.}

\begin{itemize}
    \item \texttt{"order"}: integer, starting from 1
    \item \texttt{"stage"}: short environment-appropriate stage name
    \item \texttt{"title"}: short name of the avoidance lesson
    \item \texttt{"principle"}: the concrete corrective lesson
    \item \texttt{"when\_to\_apply"}: when this lesson should be used
\end{itemize}

\textbf{Optional Fields.}

\begin{itemize}
    \item \texttt{"evidence"}:
    \texttt{"lowest\_tql\_failure"} $\mid$
    \texttt{"highest\_tql\_failure"} $\mid$
    \texttt{"mixed"}
    \item \texttt{"failure\_to\_avoid"}: common mistake this lesson prevents
\end{itemize}

\textbf{Rules.}

\begin{itemize}
    \item Summarize only stage-level lessons for this concrete task.
    \item The trajectories shown are failures; do not describe their actions as successful.
    \item Do not invent any successful outcome, unseen state, hidden observation, or successful action sequence that is not supported by the provided failed trajectories.
    \item Focus on avoidable mistakes such as misunderstanding the task goal, choosing an invalid or unhelpful action, using poor search/query/filter choices when applicable, ignoring key constraints or observations, repeating actions, acting in the wrong order, stopping too early, and missing necessary verification.
    \item When possible, identify an early mistake or recurring failure pattern that made later failure more likely, and summarize the correction at the appropriate task stage.
    \item Prefer concrete correction rules that would make the next attempt less likely to repeat the observed failures.
    \item The experience should be reusable for the same concrete task, not a verbatim copy of failed actions.
    \item If the failed trajectories do not support any reliable lesson, return [].
    \item Return only a valid JSON array and no extra text.
\end{itemize}

\end{promptbox}

\begin{promptbox}{Success-Contrast Reflection}

\textbf{Role.}

You are summarizing ordered stages for one task-level experience by comparing two successful trajectories.

\medskip
\textbf{Task Information.}

\begin{quote}
\ttfamily
Task: \{item.get("task", "")\}

Task Type: \{item.get("task\_type", "")\}

Environment/Dataset: \{env\_name\}

Selection case: success\_contrast

Missing representative types: \{json.dumps(missing, ensure\_ascii=False)\}
\end{quote}

\medskip
\textbf{Representative Trajectories.}

\begin{quote}
\ttfamily
\{self.\_format\_optional\_representative(
"BEST\_SUCCESS\_TRAJECTORY",
item.get("tau\_best\_success"))\}

\{self.\_format\_optional\_representative(
"WORST\_SUCCESS\_TRAJECTORY",
item.get("tau\_worst\_success"))\}
\end{quote}

\medskip
\textbf{Reflection Instruction.}

The worst successful trajectory is selected from baseline successful attempts, so focus on improving the base model's successful trajectory quality.

Compare two successful trajectories for the same task.

The best success has lower TQL and represents a cleaner solution.

The worst success also succeeded but has higher TQL, likely due to extra steps, repeated actions, invalid actions, or inefficient ordering.

Extract reusable task-level experience that preserves the successful strategy while avoiding inefficient successful patterns.

Produce 1 to 3 ordered TASK\_STAGE items as a JSON array. These items will be stored together as stages inside one task-level experience.

Each item must describe one distinct, reusable stage-level lesson for this concrete task.

The items must be ordered by the stage/order in which the lesson applies during task execution.

\medskip
\textbf{Required JSON Fields.}

\begin{itemize}
    \item \texttt{"order"}: integer, starting from 1
    \item \texttt{"stage"}: short environment-appropriate stage name
    \item \texttt{"title"}: short name of the experience
    \item \texttt{"principle"}: the concrete lesson
    \item \texttt{"when\_to\_apply"}: when this lesson should be used
\end{itemize}

\textbf{Optional Fields.}

\begin{itemize}
    \item \texttt{"evidence"}:
    \texttt{"best\_success"} $\mid$
    \texttt{"worst\_success"} $\mid$
    \texttt{"mixed"}
    \item \texttt{"failure\_to\_avoid"}: inefficient successful pattern this lesson prevents
\end{itemize}

\textbf{Rules.}

\begin{itemize}
    \item Summarize only stage-level lessons for this concrete task.
    \item Do not force any fixed stage schema. Use stage names appropriate to the environment and task.
    \item Preserve execution order when the trajectory shows a clear order.
    \item Do not describe WORST\_SUCCESS\_TRAJECTORY as a failure.
    \item Do not invent failure modes.
    \item Focus on efficiency, necessary ordering, redundant actions, invalid actions, repeated actions, and stage-level improvements.
    \item The experience should be reusable for the same concrete task, not a verbatim copy of actions.
    \item Return only a valid JSON array and no extra text.
\end{itemize}

\end{promptbox}

\subsection{Task-Type Experience Extraction Prompt}
\label{sec:task_type_level_prompt}

\begin{promptbox}{Task-Type-Level Experience Distillation}

\textbf{Role.}

You are abstracting task-type-level stage experiences from staged task-level experiences.

\medskip
\textbf{Task Information.}

\begin{quote}
\ttfamily
Task Type: \{task\_type\}

Environment/Dataset: \{env\_name\}
\end{quote}

\medskip
\textbf{Buffered Task-Level Experiences.}

\begin{quote}
\ttfamily
Buffered staged task-level experiences:

\{json.dumps(payload, ensure\_ascii=False, indent=2)\}
\end{quote}

\medskip
\textbf{Distillation Instruction.}

Each task-level experience describes one concrete task with task\_tql, task\_goal, and ordered stages.

Your job is to extract common reusable stage-level patterns from multiple tasks of the same task\_type.

Produce 1 to 3 task-type-level stage experiences as a JSON array.

Each item must correspond to one distinct reusable stage, decision point, or failure-avoidance pattern.

Do not create duplicate experiences.

Do not create multiple items for the same target step/stage.

Use distinct \texttt{"common\_order\_or\_stage"} values across all items.

If two candidate items would apply to the same stage, merge them into one stronger item.

\medskip
\textbf{Required JSON Fields.}

\begin{itemize}
    \item \texttt{"title"}
    \item \texttt{"principle"}
    \item \texttt{"when\_to\_apply"}
    \item \texttt{"common\_order\_or\_stage"}
    \item \texttt{"critical\_steps"}
    \item \texttt{"common\_failure\_patterns"}
    \item \texttt{"avoidance\_rules"}
\end{itemize}

\textbf{Rules.}

\begin{itemize}
    \item Extract only from the provided staged task-level experiences.
    \item Do not summarize individual tasks separately.
    \item Do not infer directly from raw trajectories except through the provided task-level experiences.
    \item Merge semantically similar stages even if objects, products, or environment details differ.
    \item \texttt{"common\_order\_or\_stage"} must be a concise semantic stage label, not a bare number or generic label such as \texttt{"1"}, \texttt{"2"}, \texttt{"stage 1"}, or \texttt{"step 2"}.
    \item Prefer patterns supported by multiple task experiences.
    \item Include a highly important single-source pattern only if it is central to the task\_type.
    \item Return only a valid JSON array and no extra text.
\end{itemize}

\end{promptbox}

\subsection{Experience Examples}

\begin{promptbox}{Example of Task-Level Experience on ALFWorld with DeepSeek-V4-Flash}

\footnotesize
\ttfamily
\raggedright
\setlength{\parskip}{0.25em}

\{

\quad "title": "Task procedure for heat some mug and put it in coffeemachine.",

\quad "principle": "Prioritize checking visible surfaces like countertops for the mug, as it is often placed there. Avoid exhaustive cabinet and drawer searches unless necessary. After obtaining the mug, go to the microwave, open it, place the mug inside, close the microwave, then execute the heat action. Omitting the heat step after closing the microwave is a critical error. After heating, take the heated mug from the microwave, go to the coffeemachine, and put the mug in it. The task requires both heating and placement; omitting the final placement leaves the task incomplete.",

\quad "when\_to\_apply": "When solving this concrete task or a very similar task: heat some mug and put it in coffeemachine.",

\quad "experience\_level": "task",

\quad "task\_type": "pick\_heat\_then\_place\_in\_recep",

\quad "source\_task": "heat some mug and put it in coffeemachine.",

\quad "retrieval\_obs": "heat some mug and put it in coffeemachine.",

\quad "stages": [

\qquad \{

\qquad\quad "order": 1,

\qquad\quad "stage": "Object Acquisition",

\qquad\quad "title": "Locate and take mug from countertop",

\qquad\quad "principle": "Prioritize checking visible surfaces like countertops for the mug, as it is often placed there. Avoid exhaustive cabinet and drawer searches unless necessary.",

\qquad\quad "when\_to\_apply": "At the start of the task, after initial observation.",

\qquad\quad "evidence": "mixed",

\qquad\quad "failure\_to\_avoid": "Unnecessary searching of multiple cabinets and drawers when the mug is already visible on a countertop."

\qquad \},

\qquad \{

\qquad\quad "order": 2,

\qquad\quad "stage": "Heating",

\qquad\quad "title": "Heat mug in microwave after placing it inside",

\qquad\quad "principle": "After obtaining the mug, go to the microwave, open it, place the mug inside, close the microwave, then execute the heat action. Omitting the heat step after closing the microwave is a critical error.",

\qquad\quad "when\_to\_apply": "After taking the mug, before moving to the coffeemachine.",

\qquad\quad "evidence": "near\_miss",

\qquad\quad "failure\_to\_avoid": "Placing mug in microwave and closing it without performing the explicit 'heat' action."

\qquad \},

\qquad \{

\qquad\quad "order": 3,

\qquad\quad "stage": "Placement",

\qquad\quad "title": "Place heated mug in coffeemachine",

\qquad\quad "principle": "After heating, take the heated mug from the microwave, go to the coffeemachine, and put the mug in it. The task requires both heating and placement; omitting the final placement leaves the task incomplete.",

\qquad\quad "when\_to\_apply": "After successfully heating the mug.",

\qquad\quad "evidence": "mixed",

\qquad\quad "failure\_to\_avoid": "Stopping after heating or becoming stuck in repetitive actions without proceeding to place the mug in the coffeemachine."

\qquad \}

\quad ],

\quad "num\_stages": 3,

\quad "stage\_when\_to\_apply": [

\qquad "At the start of the task, after initial observation.",

\qquad "After taking the mug, before moving to the coffeemachine.",

\qquad "After successfully heating the mug."

\quad ],

\quad "failure\_to\_avoid": "Unnecessary searching of multiple cabinets and drawers when the mug is already visible on a countertop.; Placing mug in microwave and closing it without performing the explicit 'heat' action.; Stopping after heating or becoming stuck in repetitive actions without proceeding to place the mug in the coffeemachine.",

\quad "env\_name": "alfworld\_parameter\_free",

\quad "data\_source": "alfworld\_parameter\_free",

\quad "utility": 0.4609896854962928,

\quad "retrieval\_count": 37,

\quad "last\_retrieval\_step": 66,

\quad "created\_at\_step": 28,

\quad "skill\_id": "task\_183"

\}

\end{promptbox}

\begin{promptbox}{Example of Task-Type-Level Experience on ALFWorld with DeepSeek-V4-Flash}

\footnotesize
\ttfamily
\raggedright
\setlength{\parskip}{0.25em}

\{

\quad "title": "Place Heated Object in Target Receptacle",

\quad "principle": "After the object is successfully heated, retrieve it from the microwave if not already in inventory. Navigate directly to the target receptacle, ensuring it is open if necessary. Use a move/put action (e.g., 'move <object> to <receptacle>') to place the object. Immediately verify that the object is no longer in inventory and appears at the receptacle. If placement failed (object still in inventory), retry the move action. Do not explore other objects or areas until placement is confirmed.",

\quad "when\_to\_apply": "Immediately after the heating action succeeds and the object is hot, before performing any other non-essential actions.",

\quad "common\_order\_or\_stage": "Final Placement",

\quad "critical\_steps": [

\qquad "If the object is still inside the microwave, open the microwave and take it out to have it in inventory",

\qquad "Navigate to the target receptacle",

\qquad "Check if the receptacle is open; if not, open it",

\qquad "Verify that the object in inventory is hot (e.g., by recent heat action or inventory state)",

\qquad "Use 'move [object] to [receptacle]' or 'put [object] on [receptacle]' to place the object",

\qquad "Check inventory or observe the receptacle to confirm placement succeeded",

\qquad "If the object remains in inventory, retry the move action"

\quad ],

\quad "common\_failure\_patterns": [

\qquad "Stopping after heating without placing the object",

\qquad "Assuming placement succeeded without verification",

\qquad "Performing unnecessary actions (examining other objects, opening containers) after heating",

\qquad "Getting stuck in loops of opening/closing the microwave or other containers",

\qquad "Incorrectly identifying the target receptacle (e.g., using wrong name)",

\qquad "Forgetting to open the receptacle before moving the object",

\qquad "Placing a wrong object (e.g., cup instead of mug)",

\qquad "Placing the object before heating (unheated placement)",

\qquad "Remaining at the microwave or wandering to other containers after heating instead of proceeding to the target receptacle"

\quad ],

\quad "avoidance\_rules": [

\qquad "After heating, immediately retrieve the object if not already in inventory",

\qquad "Navigate directly to the target receptacle without detours",

\qquad "Always ensure the receptacle is open before placing (open it if necessary)",

\qquad "Before placing, confirm that the object is hot and matches the task's exact name",

\qquad "Verify placement by checking inventory (should be empty of the object) or observing the receptacle's contents",

\qquad "If the object remains in inventory, retry the move action",

\qquad "Do not explore other areas or objects until placement is confirmed"

\quad ],

\quad "experience\_level": "type",

\quad "task\_type": "pick\_heat\_then\_place\_in\_recep",

\quad "retrieval\_obs": "Task type: pick\_heat\_then\_place\_in\_recep\textbackslash{}nStage: Final Placement\textbackslash{}nWhen: Immediately after the heating action succeeds and the object is hot, before performing any other non-essential actions.\textbackslash{}nGoal: Place Heated Object in Target Receptacle",

\quad "last\_operation": "merge",

\quad "skill\_id": "step\_020",

\quad "utility": 0.75,

\quad "retrieval\_count": 2,

\quad "last\_retrieval\_step": 18,

\quad "created\_at\_step": 7,

\quad "last\_modified\_step": 28,

\quad "version": 5,

\quad "merged\_count": 4

\}

\end{promptbox}


\end{document}